\documentclass{article} 
\usepackage{iclr2023_conference,times}

\usepackage{amsmath,amsfonts,bm}

\def\eqref#1{equation~\ref{#1}}

\def\1{\bm{1}}

\DeclareMathAlphabet{\mathsfit}{\encodingdefault}{\sfdefault}{m}{sl}
\SetMathAlphabet{\mathsfit}{bold}{\encodingdefault}{\sfdefault}{bx}{n}

\usepackage{url}
\usepackage{graphicx}

\usepackage{subfigure}
\usepackage{xfrac}
\usepackage{colortbl}
\usepackage{booktabs}  
\usepackage{amsfonts}  
\usepackage{nicefrac}
\usepackage{latexsym}
\usepackage{multirow}
\usepackage{amsmath}
\usepackage{tabularx}
\usepackage{makecell}
\usepackage{xcolor}
\usepackage{natbib}
\usepackage{wrapfig}

\usepackage{color}

\definecolor{darkblue}{rgb}{0, 0, 0.5}
\usepackage[colorlinks=true, citecolor=darkblue, linkcolor=darkblue, urlcolor=darkblue]{hyperref}  
\usepackage{paralist} 
 
\usepackage{algorithm}
\usepackage{algorithmic}

\title{Calibrating LLM-Based Evaluator}

\author{Yuxuan Liu\textsuperscript{$\dagger$}\thanks{~Work done during internship at Microsoft. \textsuperscript{$\sharp$} Corresponding author.}, Tianchi Yang\textsuperscript{$\ddagger$}, Shaohan Huang\textsuperscript{$\sharp\ddagger$}, Zihan Zhang
\textsuperscript{$\ddagger$}, Haizhen Huang\textsuperscript{$\ddagger$}, \\
\textbf{Furu Wei\textsuperscript{$\ddagger$}, Weiwei Deng\textsuperscript{$\ddagger$}, Feng Sun\textsuperscript{$\ddagger$}, Qi Zhang\textsuperscript{$\ddagger$}}
\\\\
\textsuperscript{$\dagger$} Peking University \textsuperscript{$\ddagger$} Microsoft Corporation
\\
\small\texttt{yx.liu@stu.pku.edu.cn}
}

\newcommand{\ours}{\textsc{AutoCalibrate}}
\newcommand{\ourswb}{\textsc{AutoCalibrate }}

\iclrfinalcopy 

\usepackage{tcolorbox}
\tcbuselibrary{theorems}
\newtcbtheorem[number within=section]{mytheo}{My Theorem}%
{colback=green!5,colframe=green!35!black,fonttitle=\bfseries}{th}

\begin{document}

\maketitle

\begin{abstract}
Recent advancements in large language models (LLMs) on language modeling and emergent capabilities make them a promising reference-free evaluator of natural language generation quality, and a competent alternative to human evaluation. However, hindered by the closed-source or high computational demand to host and tune,  there is a lack of practice to further calibrate an off-the-shelf LLM-based evaluator towards better human alignment. 
In this work, we propose \ours, a multi-stage, gradient-free approach to automatically calibrate and align an LLM-based evaluator toward human preference. Instead of explicitly modeling human preferences, we first implicitly encompass them within a set of human labels. Then, an initial set of scoring criteria is drafted by the language model itself, leveraging in-context learning on different few-shot examples. To further calibrate this set of criteria, we select the best performers and re-draft them with self-refinement. Our experiments on multiple text quality evaluation datasets illustrate a significant improvement in correlation with expert evaluation through calibration.  Our comprehensive qualitative analysis conveys insightful intuitions and observations on the essence of effective scoring criteria.\footnote{~Work in progress.}.
\end{abstract}

\section{Introduction}
\label{sec:intro}
The emergence of large language models is calling on a greater focus and importance on the quality of natural language generation evaluation. With the rapid improvement of language models, their goals have gone beyond simply fitting its output to a number of given samples to a broader human alignment. Traditional evaluation metrics like BLEU \citep{papineni2002bleu}, ROUGE \citep{lin2004rouge} and CIDEr \citep{vedantam2015cider} often require curated reference outputs, whose application is limited when the output space is open and diversified, and show a low correlation with human judgments \citep{freitag2022stopbleu}. While sophisticated model-based evaluators like BERTScore \citep{Zhang*2020BERTScore:} and COMET \citep{rei2020comet} yield correlation improvements, their performance is still limited by the quality of references. As a result, there is a surging demand for human-aligned, reference-free evaluators for NLG evaluations.

On this front, recent lines of research works explored leveraging state-of-the-art large language models (LLMs) as reference-free evaluators on various NLG tasks \citep{kocmi2023gemba, fu2023gptscore, wang2023chatgptgood, liu2023gpteval}.  Given that LLMs are optimized to follow human instructions \citep{ouyang2022training} as well as their state-of-the-art performance on language modeling \citep{OpenAI2023GPT4TR}, they could perform the task of evaluation when prompted accordingly. Multiple evidences show that LLMs are promising competent in evaluating instruction-tuned models like Alpaca \citep{alpaca} and Vicuna \citep{zheng2023judging}, and being a viable alternative to human expert evaluations \citep{zheng2023judging, dubois2023alpacafarm}.

Despite these promising results, emerging studies are raising concerns about the validity of LLM-based evaluators - whether LLM's underlying scoring mechanism aligns with human guidelines and preferences \citep{liu2023gpteval}. Existing LLM-based evaluators enclose the candidate text together with the evaluation task into an instruction prompt. While this paradigm succeeds in presenting the task, it elicits several unaddressed issues, including the sensitivity and bias to output space \citep{wang2023chatgptgood}, sample ordering \citep{wang2023notfair}, and prompt format \citep{zheng2023judging}. Plus, as the scoring prompt is also human-written, it may also incorporate potential bias to the LLM.

To address this issue, we study calibrating an LLM-based evaluator towards better human alignment. We start from a retrospection into existing LLM-based evaluators and uncover they suffer from \textit{\textbf{insufficient prompting}}, where the scoring guidelines are absent and only output spaces (e.g. 0-100) are provided, resulting in inconsistent and misaligned evaluations \citep{lu2023error}. We argue that such an issue could be mitigated by elucidating the scoring criteria. And by finalizing the scoring criteria, a \textbf{\textit{consensus}} could be reached between humans and the LLM, as a means of alignment.

However, it is non-trivial to obtain adequate criteria\footnote{~Results in \cite{chen2023exploring} suggest that poorly curated criteria reduce relevance with human expert scoring. Un-calibrated random criteria would introduce extra bias as a misalignment between the standards used for human experts. And improperly assigned rubrics might reduce the difference between each score.}, as it may require expert-level domain knowledge to assign rubrics and prevent personal bias. Drawing inspirations from the in-context learning capability \citep{dong2022survey} of LLMs, we propose \ours, a framework to automatically align and calibrate an LLM-based evaluator through human alignment. To tackle the challenge of curating scoring criteria, we take a data-driven methodology to draft, filter, and refine rubrics using the LLM, based on human expert labels. By incorporating the mined and calibrated rubrics into scoring instructions, we obtained significant improvements in human alignment when evaluating text summarization, data-to-text generation, and hallucinations. Moreover, we release the optimal scoring criteria sets mined for the above tasks, and present detailed qualitative and quantitative analysis to uncover the essence that makes an effective criteria.

\begin{figure*}[t]
  \centering
  \includegraphics[width=1.0\textwidth]{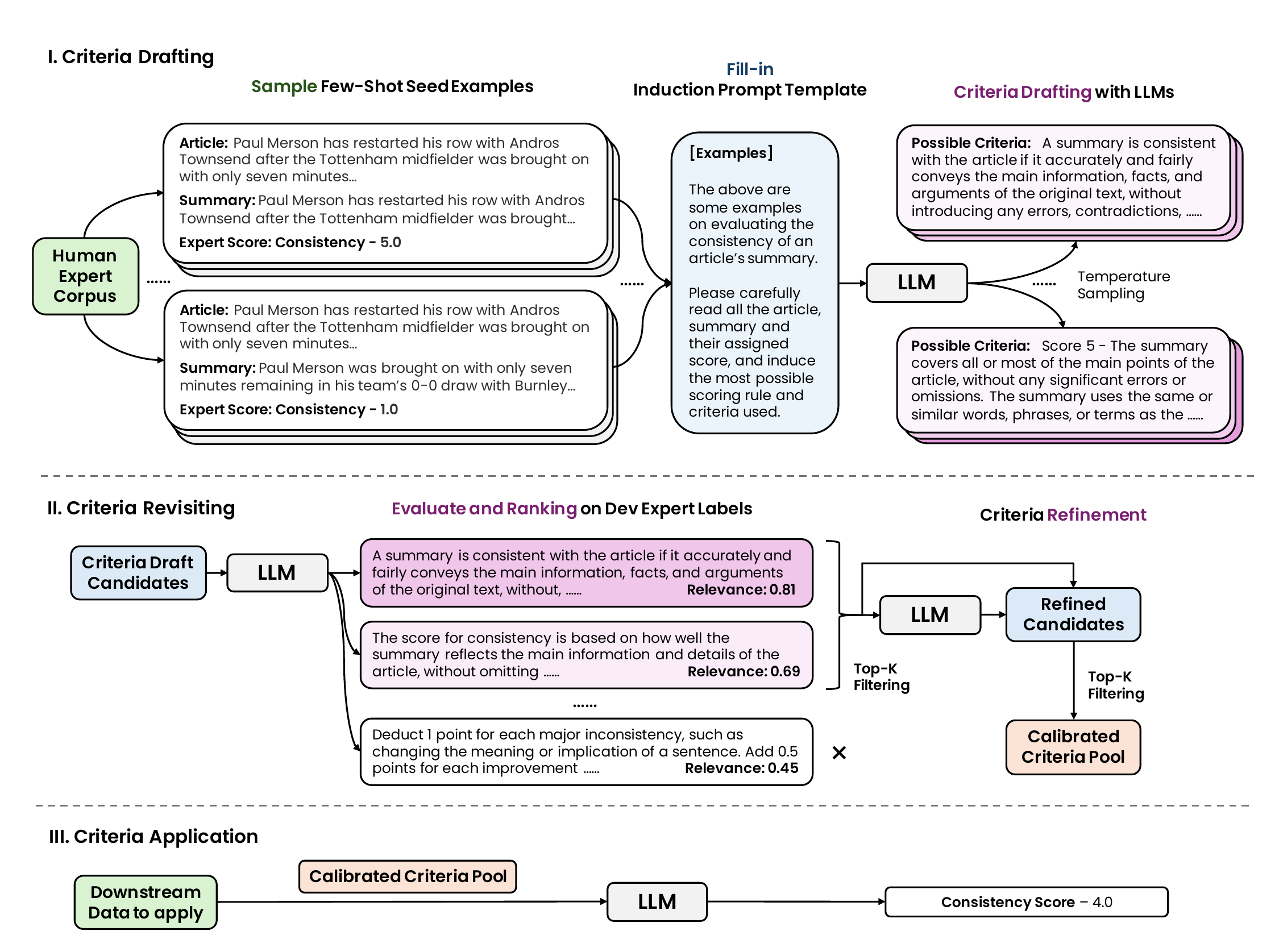}
      \caption{Overall framework of \ours. To calibrate a LLM-based evaluator towards better alignment with human expert preference, we propose a 3-stage procedure to draft, revisit, and apply high-quality scoring criteria. }
  \label{fig:overview}
\end{figure*}


\section{Methodology}
\subsection{Overview of \ours}

Figure \ref{fig:overview} illustrates the overall framework of \ours. To calibrate an LLM-based evaluator, we focus on optimizing the evaluation prompt template $\mathcal{T}$ applied to improve the correlation and alignment between LLM's scores and human preference. Specifically, we mine and tune the \textit{\textbf{scoring criteria}} in pursuing such alignment. To express human preference, we first construct a golden set $D^*$, containing ground-truth sample-label pairs from human expert labeler. We then follow a novel multi-stage procedure to optimize candidate scoring criteria, including drafting and revisiting. Initial criteria drafts are first inferred from in-context labels and an induction prompt, evaluated and filtered on expert labels, and then refined to accommodate erroneous evaluations.

\subsection{Problem formulation}
\begin{figure*}[t]
  \centering
  \includegraphics[width=0.9\textwidth]{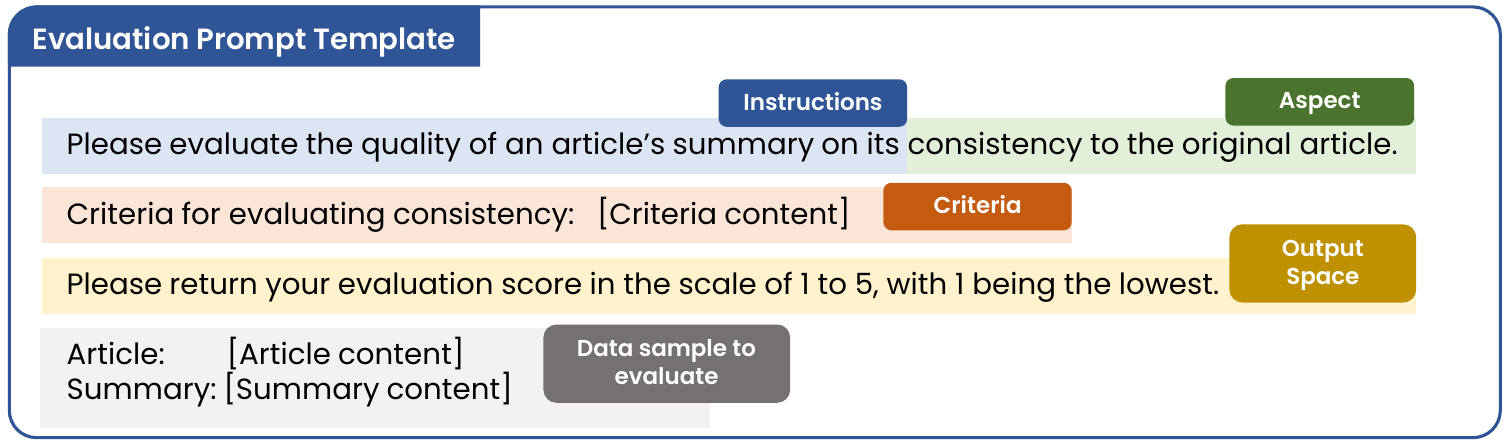}
      \caption{Example of an evaluation prompt template applied by a LLM-based evaluator.}
  \label{fig:prompttemplate}
\end{figure*}

In this section, we elaborate on the calibration medium and objective of \ourswb - the scoring criteria. Denote $D$ the dataset which contains multiple samples to evaluate. Based on different tasks, a sample $d_i \in D$ can contain various components: \textit{single text}, for tasks like evaluating grammatical correctness; \textit{source-target}, for the vast majority of conditional generations, and \textit{multi-turn}, like assessing multi-turn conversations.

To guide the LLM to evaluate the quality of sample $d_i$, prompts are applied to provide sufficient instructions and clarifications of the task. To calibrate the prompt template $\mathcal{T}$ applied during evaluation, we regulate it by decomposing it into the following building blocks: \textit{instructions, criteria, aspect, output format,} and \textit{data sample} to evaluate, as illustrated in Figure \ref{fig:prompttemplate}.
For an arbitrary sample $d_i \in D$, given a prompt template $\mathcal{T}$ (guides the LLM to perform evaluation on NLG quality), scoring criteria $\mathcal{C}$, evaluation aspect $a$ (e.g., fluency, coherence, consistency) and a large language model $LLM(\cdot)$, the NLG quality of $d_i$ could be evaluated as
\begin{equation}
\label{score}
    \hat{s}_{i,a} = LLM(\mathcal{T}(d_i, \mathcal{C}, a)).
\end{equation}
Denote $D^*$ a golden set consists of curated sample-label pairs $(d_i^*,s_{i,a})$ from human experts, and $f(\cdot)$ an correlation metric. In \ours, we focus on calibrating the scoring criteria $\mathcal{C}$ to maximize the correlation between predicted labels and human expert labels, as
\begin{equation}
    \mathcal{C} = \arg \max_\mathcal{C} f\left[\cup_{d^*_i \sim D^*}\left(\hat{s}_{i,a}, s_{i,a}\right)\right].
\end{equation}

\subsection{\ours}
\label{sec:method}
\paragraph{Data Labeling as Human Preference} To calibrate an LLM-based evaluator, one primary question is: \textit{how} to represent and model the preference of human experts. On existing approaches, sophisticated model-based evaluators like COMET \citep{rei2020comet} directly train on human labels, while ranking-based human labels are widely adopted in RLHF to model human preference \citep{ouyang2022training}. However, these model-based preference modeling methods require extra fine-tuning, which makes them computationally intensive and impracticable to API-based LLMs. To mitigate these limitations, We implicitly encode human expert preference to a set of sample-label pairs and form a golden set $D^*$. Compared with curating finalized scoring criteria and guidelines with joint human expert works, it is more feasible to collect labels leveraging crowd-sourcing dispatch, and also easier to validate and merge opinions from different experts.

\paragraph{Criteria Drafting} After constructing the expert label set $D^*$, we utilize the instruction following and in-context learning capability of LLMs to independently infer scoring criteria $\mathcal{C}$ from few-shot exemplars. One crucial part here is to ensure the diversity of recalled criteria. To mitigate the label bias and position bias of in-context learning \citep{zhao2021calibrate}, we construct various Monte-Carlo samples from $D^*$ to obtain few-shot in-context exemplars. Given drafting prompt template $\mathcal{T}_D$ and a few-shot exemplar set $D_s = \cup(d_i^*,s_{i,a}) \subset D^*$, an corresponding criteria is inferred as
\begin{equation}
    \hat{\mathcal{C}} = \arg \max_{\mathcal{C}} \mathrm{P}_\theta (\mathcal{C} | \mathcal{T}_D(D_s,a)),
    \label{eq:indu}
\end{equation}
where $a$ denotes the evaluation aspect. Temperature sampling is also applied to draw scoring criteria in diversified presentations from the LLM. Example prompt templates are provided in Appendix \ref{app:indu_templ}. Following this procedure, we obtain the initial set of scoring criteria for evaluation and refinement.

\paragraph{Criteria Revisiting}
Inferred from various few-shot exemplars, criteria within the initial draft set are diversified, but may be sub-optimal or contain potential bias (e.g., to particular scoring labels). To filter out high-quality candidates, we first revisit them leveraging $D^*$ and select the top performing candidates w.r.t their human relevance\footnote{~ A meta-evaluation method $f(\cdot)$ is applied here to perform meta-evaluation on the correlation between human and LLM judgments. For detailed explanations and definitions, please refer to Appendix \ref{app:strategy}.}. To mitigate disagreements between human experts and the drafted criteria, we prompt LLMs to refine \citep{madaan2023self} the previously generated criteria by providing them samples with strong disagreement in their scores. When refining the criteria, we suggest the following atomic editing operations via prompting to the LLM\footnote{~Detailed prompt examples are provided in Appendix \ref{app:sr_templ}.}:
\begin{itemize}
    \item \textit{Modification}: Adjust some parts of the criteria to increase its correlation.
    \item \textit{Paraphrase}: If a criteria is good enough, paraphrase it to make it clearer and more concise.
    \item \textit{Adding Aspects or Details}: When LLM discovers new underlying scoring rules that are not covered by the current criteria, consider adding them as a new line to the current criteria, but make sure not to make the criteria too long and redundant.
    \item \textit{Calibrate}: Any other modifications that the LLM considers helpful.
\end{itemize}

As illustrated in Figure \ref{fig:overview}, after obtaining refined candidate criteria, we first filter them with $D^*$ and then combine them with the pre-filtered draft criteria to obtain a calibrated set of scoring rules.

\paragraph{Conclusion and Discussion} Combining the above, we obtain \ours, an automatic pipeline in calibrating LLM-based evaluators. The overall procedure is summarized in Algorithm \ref{algo:overall}.

The benefits of choosing criteria as a medium for calibration are multitudinous. First, we do not require gradients or access to model parameters, which makes \ourswb applicable to API-based LLMs. Second, since criteria remain in a natural language form (compared with soft prompt-tuning), calibrating the criteria is essential to reaching an agreement between humans and the LLM. Therefore, the process is more explainable and controllable (e.g., one may perform \textit{human-in-the-loop} adjustments to scoring criteria in case of preference changes, or to avoid corner cases).

\begin{algorithm}[tb] 
    \caption{Calibrating LLM-Based Evaluator with \ours}\label{alg:mainalgo}
    \label{algo:overall}
    \begin{algorithmic}
        \STATE {\bfseries Require:} LLM $\theta$, human expert labels $D^*$, meta-correlation metric $f(\cdot)$, Monte-Carlo trial count $N$, in-context exemplar size $L= \{ l_1, ..., l_m \}$, aspect $a$, target criteria candidate pool size $K$
    \end{algorithmic}
    \begin{algorithmic}[1]
        \FOR{Few-shot exemplar size $l_i$ in $L$} 
            \FOR{Monte-Carlo trial $j$ in $1:N$} 
                \STATE Sample few-shot examples of human labels $D_s = \cup(d_i^*,s_{i,a})$ from $D^*$ 
                \STATE Draft candidate criteria with LLM according to Eq.(\ref{eq:indu}) using temperature sampling
                \STATE Add the obtained criteria $\mathcal{C}_i$ to candidate set $\mathcal{C}$
            \ENDFOR
        \ENDFOR
        \STATE Revisit $\mathcal{C}$ and retain top-K candidates with highest correlation: $\mathcal{C} \leftarrow \mathrm{argTopK}_{c_i \in \mathcal{C}} f(c_i,D^*)$
        \STATE Collect mis-aligned evaluation examples as $D_i^R$ for each $c_i$ in $\mathcal{C}$
        \FOR{Candidate criteria $c_i$ in $\mathcal{C}$}
            \FOR{Monte-Carlo trial $j$ in $1:N$}
                \STATE Sample few-shot examples of misaligned labels $D_s^R = \cup(d_i^R,s_{i,a}^R)$ from $D_i^R$
                \STATE Refine candidate criteria with LLM, then add the obtained criteria to candidate set $\mathcal{C}$
            \ENDFOR
        \ENDFOR
    \end{algorithmic}
    \begin{algorithmic}
        \STATE {\bfseries Return} Calibrated criteria $\mathcal{C}_{fin} \leftarrow \mathrm{argmax}_{c_i \in \mathcal{C}} f(c_i,D^*)$
    \end{algorithmic}
\end{algorithm}

\section{Experimental Setup}

\subsection{Tasks and Datasets}
We evaluate \ourswb on three text quality evaluation tasks, including\textit{ text summarization, data-to-text generation}, and \textit{evaluating hallucinations}. We select tasks following previous research works \citep{zhong2022unieval, fu2023gptscore}. We select two datasets for each of the tasks, consisting of 6 datasets in total, each containing human expert labels for candidate samples. Specifically, we select NewsRoom \citep{grusky2018newsroom} and SummEval \citep{fabbri2021summeval} for evaluating machine summarization; SFRES \citep{wen2015sfressfhot} and SFHOT \citep{wen2015sfressfhot} for data-to-text task, QAGS-XSUM and QAGS-CNN \citep{wang2020qags} for evaluating hallucinations. To evaluate the alignment between the scoring from LLM and human experts, we perform a meta-evaluation following \citep{zhong2022unieval}. Details on the evaluation strategy are listed in Appendix \ref{app:strategy}.

\subsection{Models and Baselines}
To implement \ours, we select OpenAI's GPT-4 model (\texttt{GPT-4-32K}) as the LLM for the evaluator. We list prompt templates for criteria drafting, evaluation, and refinement for tach tasks in Appendix \ref{app:prompt_list}. We set the temperature to $0$ during evaluation, and $1$ when obtaining initial criteria drafts and their refined versions. Please refer to Appendix \ref{app:impl_detail} for detailed configurations of each task.

We compare \ourswb with various state-of-the-art and/or widely applied evaluators. We first include ROUGE \citep{lin2004rouge}, a widely-applied n-gram-based evaluation metric for text summarization. We then select various evaluators based on smaller neural (language) models, including BERTScore \citep{Zhang*2020BERTScore:}, MoverScore \citep{zhao2019moverscore}, PRISM \citep{thompson2020prism}, BartScore \citep{yuan2021bartscore}, CTC \citep{deng2021CTC}, and UniEval \citep{zhong2022unieval}. Finally, we compare evaluators based on state-of-the-art LLMs (e.g. GPT-3.5 and GPT-4), including GPTScore \citep{fu2023gptscore}, ChatGPT\footnote{~The `ChatGPT' evaluator included multiple versions according to different prompt templates, and we mark these variants with parentheses. We encourage readers to check the original works for detailed information.} \citep{wang2023chatgptgood}, and GPT-Eval \citep{liu2023gpteval}. 

\section{Experimental Results}

\subsection{Results for Summarization}
We conduct meta-correlation analysis on NewsRoom and SummEval benchmark to evaluate \ours's performance to calibrate an LLM-based evaluator on text summarization. Following \cite{liu2021explainaboard}, we perform summary-level Spearman and Kendall correlation analysis on each of the 4 evaluation metrics with human expert evaluations. To represent the performance of our \textit{un-calibrated} backbone LLM, we add a \textsc{GPT-4} baseline, whose evaluations are obtained with a one-pass call using an evaluation prompt where scoring criteria is omitted \footnote{~For a fair comparison, the only difference is the removal of criteria from prompt. We keep the rest identical.}.

Results on NewsRoom and SummEval benchmark are listed in Table \ref{tab:nsrm} and \ref{tab:sme}, respectively. On NewsRoom benchmark (Table \ref{tab:nsrm}), our \ourswb significantly outperforms the LLM-based ChatGPT evaluator. It also surpasses the vanilla GPT-4-based evaluator by a large margin (with a 10.4\% improvement on Spearman and 11\% on Kendall correlation), demonstrating the importance and effectiveness of the calibration procedure. While BartScore obtained a competent performance on NewsRoom, it falls short on SummEval. We conjecture that since it utilizes a smaller model, the consistency of its scoring might be hindered due to the distribution of its fine-tuning corpus. 

In contrast, our \ourswb demonstrated a consistent human relevance uplift on both summarization datasets, since the pretraining knowledge in LLM is more in-depth and generalizable. On SummEval, \ourswb improves the human correlation of GPT-4 evaluations by 7.3\%, and also superior to a strong baseline \textsc{G-Eval-4} that also utilizes GPT-4. Noteworthy, \textsc{G-Eval-4} requires 20 calls from LLM to obtain an average score to mitigate replicated evaluations. While this improves Spearman correlation by creating a more continuous distribution, it reduces the rank coefficient. In contrast, by \textit{elucidating the scoring rule} with calibrated criteria, \ourswb improves both Spearman (2.9\%) and Kendall (13.4\%) coefficients with only one forward call.

\renewcommand\arraystretch{1.1}
\begin{table*}[t]
\center \footnotesize
\tabcolsep0.1 in
\resizebox{1.0\textwidth}{!}{

\begin{tabular}{lcccccccccc}
\toprule
\multicolumn{1}{l}{\multirow{2}[1]{*}{\textbf{Metrics}}} & \multicolumn{2}{c}{\textbf{Coherence}}
 & \multicolumn{2}{c}{\textbf{Relevance}} & \multicolumn{2}{c}{\textbf{Informative}} & \multicolumn{2}{c}{\textbf{Fluency}} & \multicolumn{2}{c}{\textbf{Average}} \\
 & \multicolumn{1}{c}{$\rho$} & \multicolumn{1}{c}{$\tau$} & \multicolumn{1}{c}{$\rho$} & \multicolumn{1}{c}{$\tau$} & \multicolumn{1}{c}{$\rho$} & \multicolumn{1}{c}{$\tau$} & \multicolumn{1}{c}{$\rho$} & \multicolumn{1}{c}{$\tau$} & \multicolumn{1}{c}{$\rho$} & \multicolumn{1}{c}{$\tau$}  \\
 
\cmidrule(lr){1-1} \cmidrule(lr){2-3} \cmidrule(lr){4-5} \cmidrule(lr){6-7} \cmidrule(lr){8-9} \cmidrule(lr){10-11}

ROUGE-1  & 0.095 & 0.076 & 0.147 & 0.112 & 0.130 & 0.099 & 0.104 & 0.082 & \cellcolor[rgb]{ .851,  .851,  .851}0.119 &\cellcolor[rgb]{ .851,  .851,  .851}0.092 \\
\textsc{ROUGE-2} & 0.026 & 0.009 & 0.091 & 0.065 & 0.079 & 0.052 & 0.048 & 0.032 & \cellcolor[rgb]{ .851,  .851,  .851}0.061 & \cellcolor[rgb]{ .851,  .851,  .851}0.092 \\
\textsc{ROUGE-L} &0.064 &0.051 &0.106 &0.083 &0.089 &0.064 &0.072 &0.061 & \cellcolor[rgb]{ .851,  .851,  .851}0.083 & \cellcolor[rgb]{ .851,  .851,  .851}0.065 \\
\textsc{BERTScore} &0.147 &0.116 &0.162 &0.126 &0.130 &0.105 &0.171 &0.128 &\cellcolor[rgb]{ .851,  .851,  .851}0.152 &\cellcolor[rgb]{ .851,  .851,  .851}0.119 \\
\textsc{MoverScore} &0.161 &0.127 &0.195 &0.157 &0.188 &0.151 &0.120 &0.086 &\cellcolor[rgb]{ .851,  .851,  .851}0.166 &\cellcolor[rgb]{ .851,  .851,  .851}0.130 \\
\textsc{PRISM} &0.573 &0.478 &0.553 &0.460 &0.561 &0.472 &0.532 &0.443 &\cellcolor[rgb]{ .851,  .851,  .851}0.555 &\cellcolor[rgb]{ .851,  .851,  .851}0.463 \\
\textsc{BARTScore} (CNN) &\textbf{0.653} &\textbf{0.547} &0.567 &0.478 &0.616 &0.510 &\textbf{0.640} &0.540 &\cellcolor[rgb]{ .851,  .851,  .851}0.619 &\cellcolor[rgb]{ .851,  .851,  .851}0.519 \\

\midrule

\textsc{ChatGPT} (DA)  &0.469 &0.405 &0.461 &0.392 &0.578 &0.498 &0.507 &0.427 &\cellcolor[rgb]{ .851,  .851,  .851}0.504 &\cellcolor[rgb]{ .851,  .851,  .851}0.430 \\

\textsc{ChatGPT} (Stars) &0.428 &0.375 &0.402 &0.348 &0.557 &0.487 &0.451 &0.385 &\cellcolor[rgb]{ .851,  .851,  .851}0.460 &\cellcolor[rgb]{ .851,  .851,  .851}0.399 \\





\midrule
\textsc{GPT-4} & 0.557 & 0.498 & 0.574 & 0.511 & 0.581 & 0.521 & 0.601 & 0.535 &\cellcolor[rgb]{ .851,  .851,  .851} 0.578 &\cellcolor[rgb]{ .851,  .851,  .851} 0.516 \\
\ours  & 0.602 & 0.540 & \textbf{0.656} & \textbf{0.585} & \textbf{0.654} & \textbf{0.590} & \textbf{0.640} & \textbf{0.575} &\cellcolor[rgb]{ .851,  .851,  .851} \textbf{0.638} &\cellcolor[rgb]{ .851,  .851,  .851} \textbf{0.573}\\
\bottomrule
\end{tabular}
}

\caption{Summary-level Spearman ($\rho$) and Kendall ($\tau$) correlations of aspects on NewsRoom.}
\label{tab:nsrm}

\end{table*}

\renewcommand\arraystretch{1.1}
\begin{table*}[t]
\center \footnotesize
\tabcolsep0.1 in
\resizebox{1.0\textwidth}{!}{

\begin{tabular}{lcccccccccc}
\toprule
\multicolumn{1}{l}{\multirow{2}[1]{*}{\textbf{Metrics}}} & \multicolumn{2}{c}{\textbf{Coherence}}
 & \multicolumn{2}{c}{\textbf{Consistency}} & \multicolumn{2}{c}{\textbf{Fluency}} & \multicolumn{2}{c}{\textbf{Relevance}} & \multicolumn{2}{c}{\textbf{Average}} \\
 & \multicolumn{1}{c}{$\rho$} & \multicolumn{1}{c}{$\tau$} & \multicolumn{1}{c}{$\rho$} & \multicolumn{1}{c}{$\tau$} & \multicolumn{1}{c}{$\rho$} & \multicolumn{1}{c}{$\tau$} & \multicolumn{1}{c}{$\rho$} & \multicolumn{1}{c}{$\tau$} & \multicolumn{1}{c}{$\rho$} & \multicolumn{1}{c}{$\tau$}  \\
 
\cmidrule(lr){1-1} \cmidrule(lr){2-3} \cmidrule(lr){4-5} \cmidrule(lr){6-7} \cmidrule(lr){8-9} \cmidrule(lr){10-11}

\textsc{ROUGE-1} & 0.167 & 0.126 & 0.160 & 0.130 & 0.115 & 0.094 & 0.326 & 0.252 & \cellcolor[rgb]{ .851,  .851,  .851} 0.192 & \cellcolor[rgb]{ .851,  .851,  .851} 0.150 \\
\textsc{ROUGE-2} & 0.184 & 0.139 & 0.187 & 0.155 & 0.159 & 0.128 & 0.290 & 0.219 & \cellcolor[rgb]{ .851,  .851,  .851} 0.205 & \cellcolor[rgb]{ .851,  .851,  .851} 0.161 \\
\textsc{ROUGE-L} & 0.128 & 0.099 & 0.115 & 0.092 & 0.105 & 0.084 & 0.311 & 0.237 & \cellcolor[rgb]{ .851,  .851,  .851} 0.165 & \cellcolor[rgb]{ .851,  .851,  .851} 0.128 \\
\textsc{BertScore} & 0.284 & 0.211 & 0.110 & 0.090 & 0.193 & 0.158 & 0.312 & 0.243 & \cellcolor[rgb]{ .851,  .851,  .851} 0.225 & \cellcolor[rgb]{ .851,  .851,  .851} 0.175 \\
\textsc{MoverScore} & 0.159 & 0.118 & 0.157 & 0.127 & 0.129 & 0.105 & 0.318 & 0.244 & \cellcolor[rgb]{ .851,  .851,  .851} 0.191 & \cellcolor[rgb]{ .851,  .851,  .851} 0.148 \\

\textsc{PRISM} &0.249 &0.196 &0.212 &0.163 &0.345 &0.285 &0.254 &0.205 &\cellcolor[rgb]{ .851,  .851,  .851} 0.265 &\cellcolor[rgb]{ .851,  .851,  .851}0.212 \\

\textsc{CTC} (Consistency) & 0.223 & 0.172 & 0.415 & 0.345 & 0.335 & 0.276 & 0.166 & 0.124 & \cellcolor[rgb]{ .851,  .851,  .851} 0.285 & \cellcolor[rgb]{ .851,  .851,  .851} 0.229 \\

\textsc{CTC} (Relevance) & 0.402 & 0.310 & 0.366 & 0.301 & 0.299 & 0.245 & 0.428 & 0.336 & \cellcolor[rgb]{ .851,  .851,  .851} 0.374 & \cellcolor[rgb]{ .851,  .851,  .851} 0.298 \\

\textsc{BartScore} (CNN) & 0.448 & 0.342 & 0.382 & 0.315 & 0.356 & 0.292 & 0.356 & 0.273 & \cellcolor[rgb]{ .851,  .851,  .851} 0.385 & \cellcolor[rgb]{ .851,  .851,  .851} 0.305 \\

\textsc{UniEval} (Multi-task) & 0.495 & 0.374 & 0.435 & 0.365 & 0.419 & 0.346 & 0.424 & 0.327 &  \cellcolor[rgb]{ .851,  .851,  .851} 0.443 & \cellcolor[rgb]{ .851,  .851,  .851} 0.353 \\

\textsc{UniEval} (Continual) & 0.575 & 0.442 & 0.446 & 0.371 & 0.449 & 0.371 & 0.426 & 0.325 & \cellcolor[rgb]{ .851,  .851,  .851} 0.474 & \cellcolor[rgb]{ .851,  .851,  .851} 0.377 \\

\midrule
\textsc{ChatGPT} (DA)  & 0.451 & 0.383 & 0.432 & 0.399 & 0.380 & 0.351 & 0.439 & 0.379 & \cellcolor[rgb]{ .851,  .851,  .851} 0.425 &  \cellcolor[rgb]{ .851,  .851,  .851} 0.378\\


\textsc{G-Eval-3.5} & 0.440 & 0.335 & 0.386 & 0.318 & 0.424 & 0.347 & 0.385 & 0.293 & \cellcolor[rgb]{ .851,  .851,  .851} 0.401 & \cellcolor[rgb]{ .851,  .851,  .851} 0.320 \\

\textsc{G-Eval-4} & \textbf{0.582} & 0.457 &\textbf {0.507} & 0.425 & 0.455 & 0.378 & 0.547 & 0.433 & \cellcolor[rgb]{ .851,  .851,  .851} 0.514 & \cellcolor[rgb]{ .851,  .851,  .851} 0.418 \\

\midrule
\textsc{GPT-4} & 0.535 & 0.464 & 0.466 & 0.432 & 0.440 & 0.413 & 0.532 & 0.465 & 
\cellcolor[rgb]{ .851,  .851,  .851} 0.493 &\cellcolor[rgb]{ .851,  .851,  .851} 0.443\\
\ours  & 0.570 & \textbf{0.493} & 0.500 & \textbf{0.467} & \textbf{0.487} & \textbf{0.452}& \textbf{0.560} &\textbf{0.483} &\cellcolor[rgb]{ .851,  .851,  .851} \textbf{0.529} &\cellcolor[rgb]{ .851,  .851,  .851} \textbf{0.474}\\
\bottomrule
\end{tabular}
}

\caption{Summary-level Spearman ($\rho$) and Kendall ($\tau$) correlations of aspects on SummEval.}
\label{tab:sme}

\end{table*}

\subsection{Results for Data-to-Text}
We consider SFRES and SFHOT datasets for evaluation of data-to-text generation task and follow \cite{fu2023gptscore} to conduct dataset-level meta-evaluation on human alignment. Results are listed in Table \ref{tab:d2t}. As illustrated in the table, \ourswb significantly outperforms the most competent trained evaluator (\textsc{UniEval}) over 30\%, and yields an over 20\% and 10\% improvement on Spearman correlation over \textsc{GPT-Score} (based on 175B-LLM GPT-3.5) and uncalibrated GPT-4 evaluator, respectively. These results suggest that the proposed procedures within \ourswb could promptly curate adequate scoring criteria for different NLG tasks and sample distributions.

\renewcommand\arraystretch{1.1}
\begin{table*}[t]
\center \footnotesize
\tabcolsep0.1 in
\resizebox{0.98\textwidth}{!}{

\begin{tabular}{lcccccccccc}
\toprule
\multicolumn{1}{l}{\multirow{2}[1]{*}{\textbf{Metrics}}} & \multicolumn{2}{c}{\textbf{SFRES-INF}}
 & \multicolumn{2}{c}{\textbf{SFRES-NAT}} & \multicolumn{2}{c}{\textbf{SFHOT-INF}} & \multicolumn{2}{c}{\textbf{SFHOT-NAT}} & \multicolumn{2}{c}{\textbf{Average}} \\
 & \multicolumn{1}{c}{$\rho$} & \multicolumn{1}{c}{$\tau$} & \multicolumn{1}{c}{$\rho$} & \multicolumn{1}{c}{$\tau$} & \multicolumn{1}{c}{$\rho$} & \multicolumn{1}{c}{$\tau$} & \multicolumn{1}{c}{$\rho$} & \multicolumn{1}{c}{$\tau$} & \multicolumn{1}{c}{$\rho$} & \multicolumn{1}{c}{$\tau$}  \\
 
\cmidrule(lr){1-1} \cmidrule(lr){2-3} \cmidrule(lr){4-5} \cmidrule(lr){6-7} \cmidrule(lr){8-9} \cmidrule(lr){10-11}

\textsc{ROUGE-1} & 0.129 & 0.098 & 0.109 & 0.081 & 0.116 & 0.089 & 0.113 & 0.084 & \cellcolor[rgb]{ .851,  .851,  .851} 0.117 & \cellcolor[rgb]{ .851,  .851,  .851} 0.088 \\
\textsc{ROUGE-2} & 0.124 & 0.094 & 0.094 & 0.069 & 0.080 & 0.061 & 0.086 & 0.064 &\cellcolor[rgb]{ .851,  .851,  .851} 0.096 &\cellcolor[rgb]{ .851,  .851,  .851} 0.072 \\
\textsc{ROUGE-L}  & 0.097 & 0.073 & 0.097 & 0.071 & 0.088 & 0.067 & 0.102 & 0.076 & \cellcolor[rgb]{ .851,  .851,  .851} 0.096 & \cellcolor[rgb]{ .851,  .851,  .851} 0.072 \\
\textsc{BertScore} & 0.156 & 0.119 & 0.138 & 0.102 & 0.135 & 0.104 & 0.126 & 0.094 & \cellcolor[rgb]{ .851,  .851,  .851} 0.172 & \cellcolor[rgb]{ .851,  .851,  .851} 0.105 \\
\textsc{MoverScore} & 0.021 & -0.016 & 0.075 & 0.056 & 0.042  & 0.033 & 0.038 & 0.029 & \cellcolor[rgb]{ .851,  .851,  .851} 0.044 & \cellcolor[rgb]{ .851,  .851,  .851} 0.026 \\ 
\textsc{BartScore} (CNN) & 0.154 & 0.117 & 0.138 & 0.101 & 0.164 & 0.126 & 0.075 & 0.055 & \cellcolor[rgb]{ .851,  .851,  .851} 0.133 & \cellcolor[rgb]{ .851,  .851,  .851} 0.100 \\


\textsc{UniEval} (Multi-task) & 0.225 & 0.169 & 0.333 & 0.247 & 0.249 & 0.191 & 0.320 & 0.238 & \cellcolor[rgb]{ .851,  .851,  .851} 0.282 & \cellcolor[rgb]{ .851,  .851,  .851} 0.211 \\

\midrule
\textsc{GPT-Score (D01)} & 0.270 & - & 0.317 & - & - & - & - & - & \cellcolor[rgb]{ .851,  .851,  .851} 0.294 & \cellcolor[rgb]{ .851,  .851,  .851} -\\
\textsc{GPT-Score (D03)} & 0.296 & - & 0.270 & - & - & - & - & - & \cellcolor[rgb]{ .851,  .851,  .851} 0.283 & \cellcolor[rgb]{ .851,  .851,  .851} -\\

\midrule
\textsc{GPT-4} & 0.283 & 0.247 & 0.389 & 0.329 & 0.315 & 0.277 & 0.389 & 0.331 & \cellcolor[rgb]{ .851,  .851,  .851} 0.344 & \cellcolor[rgb]{ .851,  .851,  .851} 0.296 \\
\ours & \textbf{0.315} & \textbf{0.272} & \textbf{0.416} & \textbf{0.351} & \textbf{0.357} & \textbf{0.313} & \textbf{0.440} & \textbf{0.383} & \cellcolor[rgb]{ .851,  .851,  .851} \textbf{0.382} & \cellcolor[rgb]{ .851,  .851,  .851} \textbf{0.330} \\

\bottomrule
\end{tabular}
}

\caption{Dataset-level Spearman ($\rho$) and Kendall ($\tau$) correlations of different evaluation aspects on SFRES and SFHOT. \textbf{-INF} and \textbf{-NAT} denote informativeness and naturalness, respectively.}
\label{tab:d2t}

\end{table*}
\begin{table*}[t]
\center \footnotesize
\tabcolsep0.1 in

\resizebox{0.98\textwidth}{!}{
\begin{tabular}{lccccccccc}
\toprule
\multicolumn{1}{l}{\multirow{2}[1]{*}{\textbf{Metrics}}} & \multicolumn{3}{c}{\textbf{QAGS-CNN/DM}}
 & \multicolumn{3}{c}{\textbf{QAGS-XSUM}} & \multicolumn{3}{c}{\textbf{Average}}  \\
 
 & $r$ & $\rho$ & $\tau$ &  $r$ & $\rho$ & $\tau$ & $r$ & $\rho$ & $\tau$ \\
 
\cmidrule(lr){1-1} \cmidrule(lr){2-4} \cmidrule(lr){5-7} \cmidrule(lr){8-10}

\textsc{ROUGE-1} & 0.338 & 0.318 & 0.248 & -0.008 & -0.049 & -0.040 & \cellcolor[rgb]{ .851,  .851,  .851} 0.165 & \cellcolor[rgb]{ .851,  .851,  .851} 0.134 & \cellcolor[rgb]{ .851,  .851,  .851} 0.104  \\

\textsc{ROUGE-2} & 0.459 & 0.418 & 0.333 & 0.097 & 0.083 & 0.068 & \cellcolor[rgb]{ .851,  .851,  .851} 0.278 & \cellcolor[rgb]{ .851,  .851,  .851} 0.250 & \cellcolor[rgb]{ .851,  .851,  .851} 0.200  \\

\textsc{ROUGE-L} & 0.357 & 0.324 & 0.254 & 0.024 & -0.011 & -0.009 & \cellcolor[rgb]{ .851,  .851,  .851} 0.190 & \cellcolor[rgb]{ .851,  .851,  .851} 0.156 & \cellcolor[rgb]{ .851,  .851,  .851} 0.122  \\

\textsc{BertScore} & 0.576 & 0.505 & 0.399 & 0.024 & 0.008 & 0.006 & \cellcolor[rgb]{ .851,  .851,  .851} 0.300 & \cellcolor[rgb]{ .851,  .851,  .851} 0.256 & \cellcolor[rgb]{ .851,  .851,  .851} 0.202  \\

\textsc{MoverScore} & 0.414 & 0.347 & 0.271 & 0.054 & 0.044 & 0.036 & \cellcolor[rgb]{ .851,  .851,  .851} 0.234 & \cellcolor[rgb]{ .851,  .851,  .851} 0.195 & \cellcolor[rgb]{ .851,  .851,  .851} 0.153  \\

\textsc{FactCC} & 0.416 & 0.484 & 0.376 & 0.297 & 0.259 & 0.212 & \cellcolor[rgb]{ .851,  .851,  .851} 0.356 & \cellcolor[rgb]{ .851,  .851,  .851} 0.371 & \cellcolor[rgb]{ .851,  .851,  .851} 0.294  \\


\textsc{BartScore} & 0.735 & 0.680 & 0.557 & 0.184 & 0.159 & 0.130 & \cellcolor[rgb]{ .851,  .851,  .851} 0.459 & \cellcolor[rgb]{ .851,  .851,  .851} 0.420 & \cellcolor[rgb]{ .851,  .851,  .851} 0.343  \\

CTC & 0.619 & 0.564 & 0.450 & 0.309 & 0.295 & 0.242 & \cellcolor[rgb]{ .851,  .851,  .851} 0.464 & \cellcolor[rgb]{ .851,  .851,  .851} 0.430 & \cellcolor[rgb]{ .851,  .851,  .851} 0.346  \\

\textsc{UniEval} & 0.682 & 0.662 & 0.532 & 0.461 & 0.488 & 0.399 & \cellcolor[rgb]{ .851,  .851,  .851} 0.571 & \cellcolor[rgb]{ .851,  .851,  .851} 0.575 & \cellcolor[rgb]{ .851,  .851,  .851} 0.465  \\

\midrule

\textsc{G-Eval-3.5} & 0.477 & 0.516 & 0.410 &  0.211 & 0.406 & 0.343 & \cellcolor[rgb]{ .851,  .851,  .851} 0.344 & \cellcolor[rgb]{ .851,  .851,  .851} 0.461 & \cellcolor[rgb]{ .851,  .851,  .851} 0.377 \\
\textsc{G-Eval-4} & 0.631 & 0.685 & 0.591 & 0.558 & 0.537 & 0.472 & \cellcolor[rgb]{ .851,  .851,  .851} 0.599 & \cellcolor[rgb]{ .851,  .851,  .851} 0.611 & \cellcolor[rgb]{ .851,  .851,  .851} 0.525 \\

\midrule

\textsc{GPT-4} & 0.605 & 0.649 & 0.606 & 0.637 & 0.637 & 0.637 & \cellcolor[rgb]{ .851,  .851,  .851} 0.621 & \cellcolor[rgb]{ .851,  .851,  .851} 0.643 & \cellcolor[rgb]{ .851,  .851,  .851} 0.622  \\

\ours & \textbf{0.740} & \textbf{0.744} & \textbf{0.663} & \textbf{0.662} & \textbf{0.662} & \textbf{0.662} & \cellcolor[rgb]{ .851,  .851,  .851} \textbf{0.701} & \cellcolor[rgb]{ .851,  .851,  .851} \textbf{0.703}  & \cellcolor[rgb]{ .851,  .851,  .851} \textbf{0.663} \\

\bottomrule
\end{tabular}
}
\caption{Dataset-level Pearson ($r$), Spearman ($\rho$) and Kendall-Tau ($\tau$) correlations on QAGS.}
\label{tab:qags}
\end{table*}

\subsection{Results for Evaluating Hallucinations}
Hallucinations are an important issue in NLG models where the output is based on fabricated, unwarranted facts or digress from a previous context, and it is becoming an increasingly important topic for trustworthy LLMs \citep{ji2023survey}. To test \ourswb on evaluating hallucinations, we select QAGS-CNNDM and QAGS-XSUM dataset and perform dataset-level meta-analysis following \cite{liu2023gpteval}. As presented in Table \ref{tab:qags}, \ourswb uplift the average Spearman correlation by 15\% over \textsc{G-Eval-4}. Noteworthy, since fine-tuned on CNN data, BartScore achieves promising human relevance on QAGS-CNN, but significantly falls short on QAGS-XSUM, while LLM-based \ourswb performs consistently on both datasets. This further indicates that LLMs, given their immense knowledge gained during pre-training, are strong candidates for general evaluators, and their performance could be further boosted with proper calibration.

\subsection{Ablation Experiments}
We conduct ablation studies on the procedure of \ourswb to better investigate the contribution of each process in calibrating LLM-based evaluator. The main ablation experiments are listed in Table \ref{tab:aba}. As illustrated in the table, removing criteria in the prompt significantly reduces the human correlation of GPT-4. This corroborates our argument that previously LLMs suffered from a vaguely defined scoring principle, and this could be calibrated to increase the human alignment of LLM evaluators. The self-refine process also positively contributed to the improvements in human alignment. This indicates that LLMs could accordingly adjust the effectiveness of scoring criteria. Detailed qualitative analysis is presented in Chapter \ref{ch:analysis}.


\begin{table}[]
\centering
\footnotesize
\begin{tabular}{llcccccc}
\toprule
\multicolumn{2}{l}{\multirow{2}{*}{\textbf{Dataset}}} & \multicolumn{3}{c}{$\rho$} & \multicolumn{3}{c}{$\tau$} \\
\multicolumn{2}{l}{} & OG & -Crit & -Rfi & \textbf{OG} & -Crit & -Rfi  \\ 

\cmidrule(lr){1-2} \cmidrule(lr){3-5} \cmidrule(lr){6-8}

 
\multirow{4}{*}{\begin{tabular}[c]{@{}l@{}}\textbf{News}\\ \textbf{Room}\end{tabular}} & Coherence & \textbf{0.602} & 0.557 & 0.593 & \textbf{0.540} & 0.498  & 0.531 \\
 & Relevance & \textbf{0.656} & 0.574 & 0.619  & \textbf{0.585}  & 0.511 & 0.550  \\
 & Informative & \textbf{0.654} & 0.581 & 0.617 &  \textbf{0.590} & 0.521  & 0.557 \\
 & Fluency & \textbf{0.640} & 0.601 & 0.628 & \textbf{0.575} & 0.535 & 0.563 \\ \cmidrule(lr){1-8}
 
\multirow{2}{*}{\textbf{SFRES}} & Informative & \textbf{0.315} & 0.283  & 0.300 & \textbf{0.272} & 0.247 & 0.264 \\
 & Naturalness & \textbf{0.416} & 0.389 & 0.405 & \textbf{0.351}  & 0.329 & 0.346 \\ \cmidrule(lr){1-8}
 
\multirow{2}{*}{\textbf{SFHOT}} & Informative & \textbf{0.357} & 0.315 & 0.345 &  \textbf{0.313} & 0.277 & 0.303  \\
 & Naturalness & \textbf{0.440} & 0.389 & 0.425 & \textbf{0.383} & 0.331 & 0.368  \\ \cmidrule(lr){1-8}

\multirow{2}{*}{\textbf{QAGS}} & CNN Data & \textbf{0.744} & 0.649 & 0.724 & \textbf{0.663} & 0.606 &  0.642  \\
 & XSUM Data & \textbf{0.662} & 0.637 & 0.651  & \textbf{0.662} & 0.637 & 0.651 \\ \cmidrule(lr){1-8}

\end{tabular}
\caption{Ablations on each proposed module. We report Spearman ($\rho$) and Kendall ($\tau$) correlations. `OG' denotes original method, `-Crit' and `-Rfi' denote removing criteria and refine, respectively.}
\label{tab:aba}
\end{table}

\section{Analysis}
\label{ch:analysis}

\subsection{Essence of Effective Criteria}
\begin{figure*}[t]
  \centering
  \includegraphics[width=1.0\textwidth]{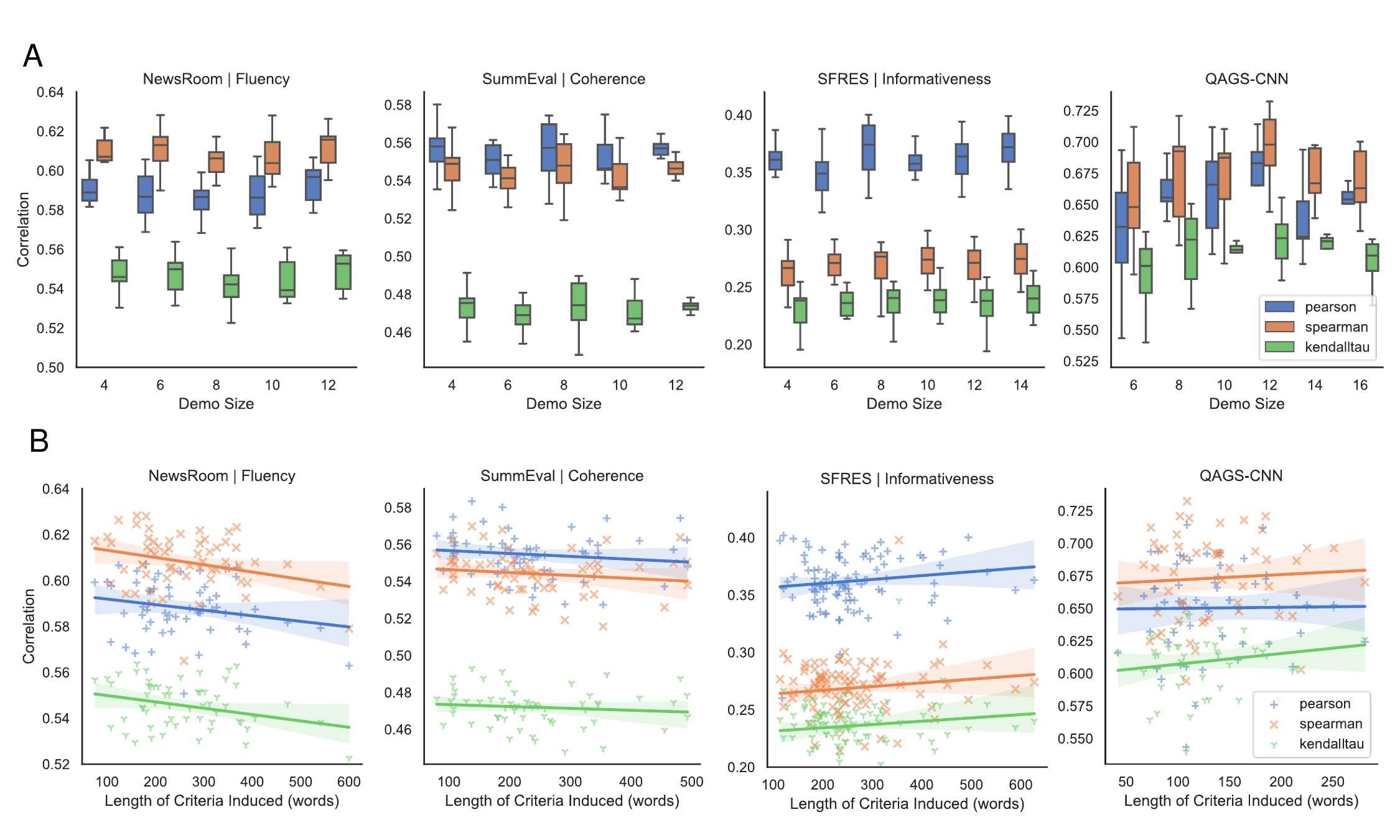}
      \caption{\textbf{Statistics of criteria} induced from \ours. A) Human correlation of criteria induced using various few-shot in-context demonstration sizes. B) Correlation between human relevance and criteria length. Shaded areas denote $95\%$ confidence interval.}
  \label{fig:compo}
\end{figure*}

In this chapter, we present statistical analysis on the pool of draft candidates of scoring criteria, and mine for possible essence that contributes to effective scoring criteria with high human relevance for LLM-based evaluators. The main results are presented in Figure \ref{fig:compo}.

\paragraph{Effect of Few-Shot Example Size} We study the sensitivity of \ourswb to the sample size of few-shot in-context samplers. As illustrated in Figure \ref{fig:compo}(A), the size of in-context few-shot exemplars yields no significant impact except for QAGS-CNN. The results indicate that \ourswb is mostly robust to the size of in-context samples. Thanks to the sufficient prior knowledge obtained during pretraining by the LLM, \ourswb is capable of inferring the underlying criteria using only a few examples in context. As illustrated in the figure, a few-shot size of 8 to 12 is sufficient in mining effective criteria across all tasks. This intriguing feature enables a reduction in search space for cost reductions upon deployment.

\paragraph{Effect of Criteria Length} The distribution of lengths of generated criteria and their human relevance is illustrated in Figure \ref{fig:compo}(B). Most evaluation criteria drafted and refined with \ourswb lie in the range of 60 to 600 words. We discover different trends on the preference of \ourswb to different lengths of criteria. While fluency and coherence metrics on text summarization lean towards shorter criteria, lengthier versions are favored by the informativeness metric on data-to-text and evaluating hallucinations. Despite this difference, \ourswb enjoys the capability to generate effective criteria at each length. We conjecture this nuance is caused by the intrinsic complexity of the aspect to evaluate: it could be straightforward to define fluency, but possibly more challenging to address hallucination. 

\begin{figure*}[t]
  \centering
  \includegraphics[width=1.0\textwidth]{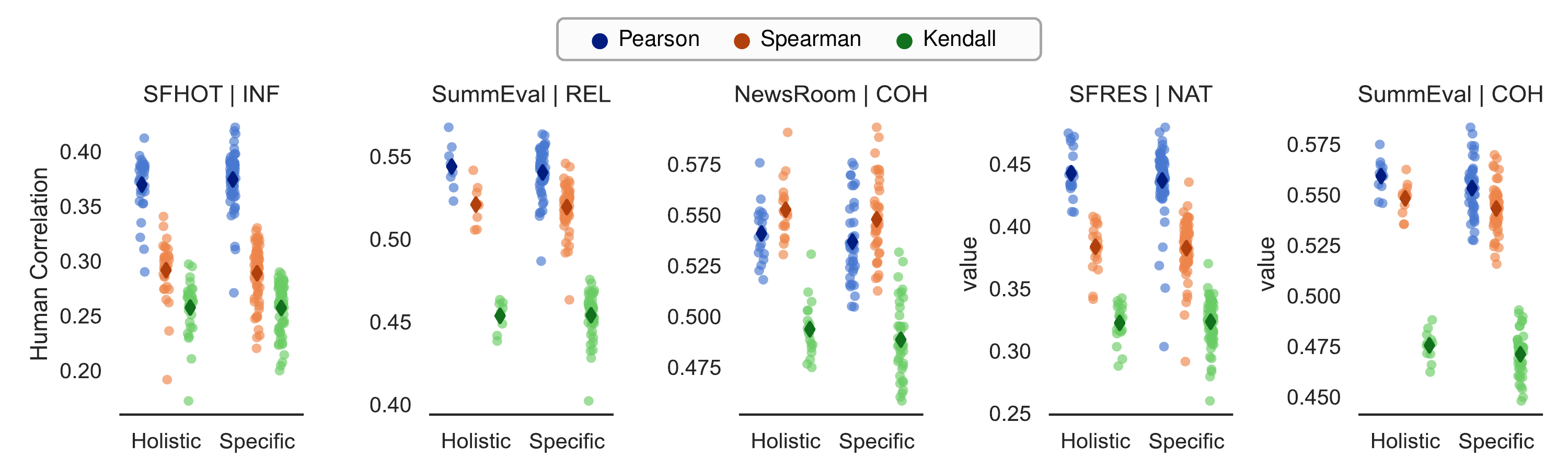}
  \vspace{-10px}
  \caption{\textbf{Performance of different patterns} of scoring criteria induced by \ours.}
  \label{fig:pattern}
\end{figure*}

\paragraph{Patterns of Criteria} We observed two significant patterns on the criteria drafted by GPT-4: \textit{holistic} and \textit{specific}. The former typically characterizes the common features possessed by high and low-quality samples, while the latter generates a segment of the corresponding rubric for each evaluation score (e.g., 1 to 5). A random example of these patterns of criteria is listed in Table \ref{tab:case-pattern}. These two patterns emerge across all sets of experiments on different benchmarks. The performance distribution of these two patterns across all datasets is illustrated in Figure \ref{fig:pattern}. As illustrated in the figure, there is no significant difference in human expert correlation between holistic and specific patterns, indicating that both patterns generated from \ourswb are of high quality. Therefore, the performance of \ourswb is robust to the patterns of criteria generated. 

\begin{table*}[t]
\center \footnotesize
\tabcolsep0.1 in

\resizebox{1\textwidth}{!}{
\begin{tabular}{p{0.9\textwidth}ccc}
\toprule
\multicolumn{1}{l}{\multirow{2}[1]{*}{\textbf{Evaluation Criteria Induced}}} & \multicolumn{3}{c}{\textbf{Human Alignment}} \\
 
 & $r$ & $\rho$ & $\tau$  \\
 
\cmidrule(lr){1-1} \cmidrule(lr){2-4}

\colorbox{green}{A} A summary should capture the main idea and key details of the article, without introducing any new or misleading information. A summary should use clear and concise language, avoiding unnecessary repetition or filler words. A summary should be proportionate to the length and complexity of the article, reflecting the most important aspects and leaving out less relevant details.  & 0.58 & 0.56 & 0.49 \\

\midrule

\colorbox{yellow}{B} Possible scoring rule: A score of 5 means the summary is very relevant, covering all the essential elements of the article and omitting any unnecessary or misleading information. A score of 4 means the summary is mostly relevant, covering most of the essential elements of the article and omitting or including only minor or trivial information. A score of 3 means the summary is somewhat relevant, covering some of the essential elements of the article, but omitting or including some important or relevant information. A score of 2 means the summary is slightly relevant, covering only a few of the essential elements of the article, and omitting or including a lot of important or relevant information. A score of 1 means the summary is irrelevant, covering none or almost none of the essential elements of the article, and omitting or including a lot of inaccurate or irrelevant information.  & 0.53 & 0.52 & 0.45 \\

\bottomrule
\end{tabular}
}

\caption{\textbf{Case study on patterns} of criteria induced from SummEval-REL. Criteria mined tend to follow two major patterns of its form: \textit{holistic} (A) and \textit{specific} (B). The former commonly describe what makes a good or bad sample, while the latter generate specific rubrics for each of the scores.}
\label{tab:case-pattern}
\end{table*}

\subsection{Case Study}
To investigate the effect of criteria refinement, we present a case study in Table \ref{tab:case-sr}. As demonstrated in the table, when prompted with previous misaligned evaluation cases and possible means of modifications (Section \ref{sec:method}), the LLM automatically infers new patterns of underlying scoring principles, and promptly adapts the existing criteria to accommodate them. As illustrated in the table, \ourswb discovers that the \textit{genre and format} is crucial to the \textit{fluency} of summary from in-context examples provided, adjusts the criteria accordingly, and achieves higher human relevance. These findings corroborate with \cite{madaan2023self} that LLM is capable of self-refine, and opens a future research direction on the multi-turn, iterative calibration of LLM-based evaluators.
\begin{table*}[t]
\center \footnotesize
\tabcolsep0.1 in

\resizebox{1\textwidth}{!}{
\begin{tabular}{p{0.9\textwidth}ccc}
\toprule
\multicolumn{1}{l}{\multirow{2}[1]{*}{\textbf{Evaluation Criteria Induced}}} & \multicolumn{3}{c}{\textbf{Human Alignment}} \\
 & $r$ & $\rho$ & $\tau$  \\ \cmidrule(lr){1-1} \cmidrule(lr){2-4}

\colorbox{green}{Before} ... It should use appropriate vocabulary and punctuation, and avoid repetition or redundancy. It should also capture the tone and style of the original article. A summary with a medium score (3) should have few or minor errors that do not interfere with the overall meaning and readability of the summary. It should use mostly appropriate vocabulary and punctuation, and avoid repetition or redundancy. It should also capture the tone and style of the article. - A summary with a low score (1 or 2) should have frequent or major errors that affect the overall meaning and readability of the summary ... It should also fail to capture the tone and style of the original article. 
& 0.63 & 0.62 & 0.56 \\ \midrule

\colorbox{yellow}{After} ... It should also capture the tone and style of the original article and \textcolor{blue}{use the correct genre and format (e.g., not writing a summary as a list of bullet points).} A summary with a medium score (3) should have few or minor errors that do not interfere with the overall meaning and readability of the summary. It should use mostly appropriate vocabulary and punctuation, and minimize repetition or redundancy. It should also attempt to capture the tone and style of the original article \textcolor{blue}{and use the correct genre and format, but may have some inconsistencies or inaccuracies.} - A summary with a low score (1 or 2) should have frequent or major errors that affect the overall meaning and readability of the summary ... It should also fail to capture the tone and style of the original article and \textcolor{blue}{use the wrong genre or format.}
& \textbf{0.66}  & \textbf{0.64} & \textbf{0.58} \\

\bottomrule
\end{tabular}
}

\caption{\textbf{Case study of criteria refinement} on NewsRoom-FLU. To refine a criteria, the model automatically infer new patterns from bad cases and promptly adjust the criteria to incorporate them. Modifications are highlighted in \textcolor{blue}{blue}, and some parts of generated criteria are omitted for space.}
\label{tab:case-sr}
\end{table*}

\section{Related Work}
\paragraph{Automatic NLG Evaluation} It has been a long and arduous endeavor to automatically evaluate natural language generations. This paragraph outlines automatic evaluation metrics before the era of LLM. 
(1) \textit{N-gram-based metrics}: as the most widely adopted method, n-gram-based metrics measure the quality of a candidate text by the overlap of its lexical fraction between references. 
As two of the most widely used metrics, BLEU \citep{papineni2002bleu} and ROUGE \citep{lin2004rouge} are specialized in precision for machine translation and recall for text summarization, respectively. Despite being widely applied, their human relevance is undesired \citep{freitag2022stopbleu}. (2) \textit{Embedding-based metrics}: this line of method leverages a pre-trained language model (e.g. BERT \citep{bert}) to measure the similarity between word embedding of the candidate and reference text \citep{Zhang*2020BERTScore:, zhao2019moverscore}. Their major limitation lies in the similarity-based paradigm and high dependency on the quality and diversity of references. (3) \textit{Trained neural evaluators}: more recent research focus on specializing the PLMs by either fine-tuning on human \citep{rei2020comet} or synthetic \citep{zhong2022unieval} labels, or pretraining on domain-relevant documents \citep{yuan2021bartscore}. However, these metrics either focus on a single dimension \citep{wang2020asking, huang2020grade} or are limited in human relevance \citep{mehri2020usr, zhong2022unieval}.

\paragraph{LLM-Based NLG Evaluation} With the emergence of LLM, recent research works focus on LLM-based evaluators given their promising instruction-following and generalization capability. A first line of work goes through preliminary explorations on LLM-based evaluators, including prompting methods and model variants \citep{fu2023gptscore, kocmi2023gemba, wang2023chatgptgood, chen2023exploring, liu2023gpteval}. Successor research focuses on various aspects of improving LLM-based evaluators, including factuality \citep{min2023factscore}, interpretability \citep{lu2023error}, mitigating position bias \citep{wang2023notfair}, and agreement to human evaluation \citep{zheng2023judging}. Different from the above approaches, we focus on a general method to calibrate an off-the-shelf LLM with gradient-free approaches, to improve its alignment with human preferences on a desired task.

\section{Conclusion}
In this work, we focus on an important question: how to calibrate and align an off-the-shelf LLM-based evaluator towards human alignment in a gradient-free fashion. We first take a retrospection into existing LLM-based NLG evaluators and uncover they suffer from insufficient prompting, where the scoring guidelines are absent and only output spaces are provided, resulting in inconsistent and misaligned evaluations. We emphasize the significance of aligned scoring criteria as a consensus between humans and LLM and propose \ourswb to automatically calibrate an LLM-based evaluator through criteria drafting and refinement. Inferred from human expert labels and refined according to previous misalignment samples by the LLM, the criteria curated by \ourswb demonstrate significant improvements in human correlation across evaluating text summarization, data-to-text, and hallucinations. Our qualitative analysis conveys insightful intuitions and observations on the essence of effective scoring criteria.


\section{Discussions}
\paragraph{Limitations and Broader Impacts}
This work study on calibrating a strong LLM-based evaluator towards better human alignment. Beyond manual prompt engineering, \ourswb automates the calibration process of LLM-based evaluators and provides a first experimental study on how further LLM-based evaluators could be strengthened with better prompting. We envision \ourswb being potentially applied to a wider spectrum of tasks in NLG and beyond. 

The primary limitation is that only criteria are mined to improve alignment. After carefully analyzing prompts, we conclude that the criteria are most crucial, as they are most causal to the scores given, and can be regarded as a shared consensus between humans and LLMs due to their natural language form. Plus, the criteria section is the hardest to curate compared with other parts of the prompt template (e.g., scoring scale, task definition), on which we primarily focus. 
Besides, A more comprehensive research on advancing and assessing other components of prompts to calibrate a LLM-based evaluator, and adapting it to wider tasks and languages is open to future work. 




\newpage
\bibliography{main}
\bibliographystyle{iclr2023_conference}
\appendix

\section{Evaluation Strategy}
\label{app:strategy}
In this section, we introduce meta-evaluation strategies for assessing human alignment that are applied in this work. We select evaluation strategies primarily following previous works \citep{zhong2022unieval, fu2023gptscore, liu2023gpteval}. Given a dataset $\mathcal{D}$ consisting of NLG samples from $M$ diverse systems and $J$ source text samples, evaluation metric $f(\cdot)$ (e.g., BLEU \citep{papineni2002bleu}) and correlation metric $g(\cdot)$, we could perform meta-evaluation at either sample or dataset level.

\paragraph{Sample Level} For sample-level meta-evaluation, we first compute correlation values on multiple candidate response (from each system) to a individual sample, then average across all samples:
\begin{equation}
    f_{sample} = \frac{1}{J} \sum_{i=1}^{J} \left( g\left(\left[\hat{s}_{i,1}, ..., \hat{s}_{i,M}\right], \left[s_{i,1}, ..., s_{i,M}\right] \right) \right),
\end{equation}
where $\hat{s}_{u,v}$ and $s_{u,v}$ denote the evaluation results (if not, converted to a numeric value) for the $v$-th response to $u$-th sample from evaluator $f(\cdot)$ and human experts, respectively. 

\paragraph{Dataset Level} For dataset-level meta-evaluation, we evaluate the correlations on all samples in the dataset (with a total of $M \times J$ samples), as follows:
\begin{equation}
    f_{dataset} = g\left(\left[\hat{s}_{i,1}, ..., \hat{s}_{J,M}\right], \left[s_{i,1}, ..., s_{J,M}\right] \right).
\end{equation}


\section{On Performance of Adding Chain-of-Thoughts}
\label{app:cot}

\renewcommand\arraystretch{1.1}
\begin{table*}[th]
\center \footnotesize
\tabcolsep0.1 in
\resizebox{1.0\textwidth}{!}{

\begin{tabular}{lcccccccccc}
\toprule
\multicolumn{1}{l}{\multirow{2}[1]{*}{\textbf{Metrics}}} & \multicolumn{2}{c}{\textbf{Coherence}}
 & \multicolumn{2}{c}{\textbf{Consistency}} & \multicolumn{2}{c}{\textbf{Fluency}} & \multicolumn{2}{c}{\textbf{Relevance}} & \multicolumn{2}{c}{\textbf{Average}} \\
 & \multicolumn{1}{c}{$\rho$} & \multicolumn{1}{c}{$\tau$} & \multicolumn{1}{c}{$\rho$} & \multicolumn{1}{c}{$\tau$} & \multicolumn{1}{c}{$\rho$} & \multicolumn{1}{c}{$\tau$} & \multicolumn{1}{c}{$\rho$} & \multicolumn{1}{c}{$\tau$} & \multicolumn{1}{c}{$\rho$} & \multicolumn{1}{c}{$\tau$}  \\
 
\cmidrule(lr){1-1} \cmidrule(lr){2-3} \cmidrule(lr){4-5} \cmidrule(lr){6-7} \cmidrule(lr){8-9} \cmidrule(lr){10-11}

w/ Auto CoT& 0.550 & 0.477 & 0.495 & 0.461 & 0.482 & 0.447 & 0.564 & 0.492 & \cellcolor[rgb]{ .851,  .851,  .851} 0.523 & \cellcolor[rgb]{ .851,  .851,  .851} 0.469\\
w/o Auto CoT & 0.550 & 0.478 & 0.500 & 0.467 & 0.480 & 0.447& 0.560 & 0.483 & \cellcolor[rgb]{ .851,  .851,  .851} 0.523 & \cellcolor[rgb]{ .851,  .851,  .851} 0.467\\

\bottomrule
\end{tabular}
}

\caption{Performance comparison of w/ and w/o CoT in base prompt template on SummEval.}
\label{tab:cot}

\end{table*}


Chain-of-thought (CoT) \citep{wei2022chain} prompting elicits reasoning in large language models by encouraging models to generate their rationales before obtaining an answer. As studied in recent research \citep{liu2023gpteval}, chain-of-thoughts are beneficial to improving human alignment in NLG evaluation, if incorporated in the scoring prompt template. Therefore, we study whether \ourswb could further benefit from adding a CoT into our calibrated scoring prompts.

To obtain the CoT for each scoring aspect, we follow \cite{liu2023gpteval}, and results are illustrated in Table \ref{tab:cot}. As shown in the figure, adding CoTs to our calibrated prompts yields negligible difference. We conjecture the effectiveness of `CoT' is marginalized by providing informative and instructive scoring criteria. In contrast to math, the assessment of text quality is not a strictly chained reasoning process, so providing a CoT is essentially clarifying the evaluation rubrics, which is consistent with the meaning of the criteria in this paper, and thus obtained no additional benefit. Plausibly, the `CoT's here act to elucidate the scoring rules, rather than providing reasoning paths to follow.

\section{Configuration Details}
\label{app:impl_detail}
In this section, we list the configuration details of \ourswb for each experiments. Detailed configurations for \ourswb are listed in Table \ref{tab:configall}. We apply the same set of configurations to each of the two datasets within a task.

\begin{table}[th]
\centering
\footnotesize
\begin{tabular}{lccc}
\toprule
\textbf{Task}  & \textbf{Summarization} & \textbf{Data-to-text} & \textbf{Hallucination} \\ \cmidrule(lr){1-4}
Model & \textsc{GPT-4-32K} & \textsc{GPT-4-32K} & \textsc{GPT-4-32K} \\
Evaluation Temperature & 0.0 & 0.0 & 0.0 \\
Max Tokens & 20 & 20 & 20 \\ \cmidrule(lr){1-4}
Criteria Drafting Temperature & 1.0 & 1.0 & 1.0 \\
In-context Sample Size & 4,6,8,10,12 & 4,6,8,10,12,14 & 6,8,10,12,14,16 \\
Monte-Carlo Trials & 4 & 4 & 3 \\
Temperature Sampling Count & 3 & 3 & 3 \\
Max Tokens & 768 & 768 & 768 \\ \cmidrule(lr){1-4}
Criteria Refining Temperature & 1.0 & 1.0 & 1.0 \\
In-context Sample Size & 1,2,4 & 1,2,4 & 1,2,4 \\
Monte-Carlo Trials & 4 & 4 & 4 \\
Temperature Sampling Count & 2 & 2 & 2 \\ 
Max Tokens & 768 & 768 & 768 \\ \bottomrule
\end{tabular}
\caption{Detailed configurations of \ourswb for different experiments.}
\label{tab:configall}
\end{table}

\section{List of Prompt Templates}
\label{app:prompt_list}
In this section, we list prompt templates applied throughout this study, including induction templates for criteria drafting, evaluation templates that utilize the generated scoring criteria, and templates for self-refinement of criteria.

\subsection{Criteria Drafting Templates}
\label{app:indu_templ}
Prompt templates for criteria drafting are listed in Figure \ref{fig:indup-summ}, \ref{fig:indup-d2t} and \ref{fig:indup-qags}. The \textcolor{purple}{[Aspect]} denote placeholders for aspects to evaluate (e.g. coherence, consistency, etc.), and sampled few-shot in-context exemplars are placed at \textcolor{darkblue}{[In-Context Few-Shot Samples]}, including samples and their expert scores.

\begin{figure}[!t]
\centering
\begin{tcolorbox}[width=1\textwidth, fontupper=\small, colback=blue!2, boxrule=0.9pt] \#\# Instructions

Please infer the scoring criteria for the following task:

[Score the following summary of a news article on its \textcolor{purple}{[Aspect]}. Please return your score on how the summary is consistent with the article in the scale of 1 to 5, with 1 being the lowest.] \\

- The following is some examples on evaluation scores of \textcolor{purple}{[Aspect]} of summary (in the range of 1 to 5, where 1 being the lowest).

- Please carefully read all the article, summary and their assigned score, and induce the most possible scoring rule and criteria used.

- It is optimal that, by using the induced criteria, you are very likely to assign a same score on \textcolor{purple}{[Aspect]} to the provided reference scores. \\

\#\# Criteria for \textcolor{purple}{[Aspect]}

- The scoring criteria been used. Now it is not explicitly provided, and you should induce it from the following samples.

- The induced criteria should be able to explain the scores of all the samples provided, being generic and concise. \\

\#\# Examples

\textcolor{blue}{[In-Context Few-Shot Samples]} \\

\#\# Induced Criteria

Criteria for \textcolor{purple}{[Aspect]}:
\end{tcolorbox}
\caption{Prompt template for criteria drafting on text summarization (SummEval, NewsRoom).}
\label{fig:indup-summ}
\end{figure}

\begin{figure}[!t]
\centering
\begin{tcolorbox}[width=1\textwidth, fontupper=\small, colback=blue!2, boxrule=0.9pt] \#\# Instructions

Please infer the scoring criteria for the following task:

[Task data-to-text is to generate natural language sentences from structured data sources. This can be useful for creating chatbots, voice assistants, or text summarizers. Please score the following natural language sentence generated according to a structured data expression. Please return your score on \textcolor{purple}{[Aspect]} of the sentence, in the scale of 1 to 6, with 1 being the lowest.] \\

- The following is some examples on evaluation of \textcolor{purple}{[Aspect]} of the natural language sentence generated from structured data expression (in the range of 1 to 6, where 1 being the lowest).

- Please carefully read all expressions, generated sentence and its assigned score, and induce the most possible scoring rule and criteria used.

- It is optimal that, by using the same criteria, you are very likely to assign a same score to the provided reference scores.\\

\#\# Criteria for \textcolor{purple}{[Aspect]}

- The scoring criteria been used. Now it is not explicitly provided, and you should induce it from the following samples.

- The induced criteria should be able to explain the scores of all the samples provided, being generic and concise. \\

\#\# Examples

\textcolor{blue}{[In-Context Few-Shot Sampls]} \\

\#\# Induced Criteria

Criteria for \textcolor{purple}{[Aspect]}:
\end{tcolorbox}
\caption{Prompt template for criteria drafting on data-to-text (SFRES, SFHOT).}
\label{fig:indup-d2t}
\end{figure}

\begin{figure}[!t]
\centering
\begin{tcolorbox}[width=1\textwidth, fontupper=\small, colback=blue!2, boxrule=0.9pt] \#\# Instructions

Please infer the scoring criteria for the following task:

[Evaluate the \textcolor{purple}{factual consistency} of the summary to the article. Check how well the summary is supported by the article and whether it contains untruthful or misleading facts. Score 1 if the summary is factually consistent with the article, 0 otherwise.] \\

- The following is some examples on evaluation of \textcolor{purple}{factual consistency} of generated summary to the article.

- Please carefully read all summary - article pairs and its assigned score, and induce the most possible scoring rule and criteria used.

- It is optimal that, by using the same criteria, you are very likely to assign a same score to the provided reference scores. \\

\#\# Criteria for \textcolor{purple}{factual consistency}

- The scoring criteria been used. Now it is not explicitly provided, and you should induce it from the following samples.

- The induced criteria should be able to explain the scores of all the samples provided, being generic and concise. \\

\#\# Examples

\textcolor{blue}{[In-Context Few-Shot Samples]} \\

\#\# Induced Criteria

Criteria for \textcolor{purple}{factual consistency}:
\end{tcolorbox}
\caption{Prompt template for criteria drafting on evaluating hallucinations (QAGS-XSUM/CNN).}
\label{fig:indup-qags}
\end{figure}

\subsection{Evaluation Templates}
\label{app:eval_templ}
Prompt templates for evaluation are listed in Figure \ref{fig:eval-summ}, \ref{fig:eval-d2t} and \ref{fig:eval-qags}. The \textcolor{purple}{[Aspect]} denotes placeholders for aspects to evaluate (e.g. coherence, consistency, etc.). Evaluation samples and calibrated scoring criteria for each aspect are filled into corresponding placeholders during evaluation. 

\begin{figure}[!t]
\centering
\begin{tcolorbox}[width=1\textwidth, fontupper=\small, colback=blue!2, boxrule=0.9pt] 
\#\# Instructions

Score the following summary of a news article on its \textcolor{purple}{[Aspect]}.

Please return your score on how the summary is \textcolor{purple}{[Aspect]} with the article in the scale of 1 to 5, with 1 being the lowest.\\

\#\# Example

\textcolor{blue}{[Article and Summary to be evaluated]} \\

\#\# Criteria for \textcolor{purple}{[Aspect]}

\textcolor{blue}{[Calibrated criteria for evaluating this aspect]} \\

\#\# Evaluation

Now, please evaluate how \textcolor{purple}{[Aspect]} is the summary to the article (on a scale of 1-5, with 1 being the lowest).
Please carefully read the article and summary, and follow the scoring criteria above to score the \textcolor{purple}{[Aspect]} of the summary to the article.

Please first return your score, and then provide your reasoning for the score. \\

\textcolor{purple}{[Aspect]} Score (1-5):
\end{tcolorbox}
\caption{Prompt template for evaluation on text summarization (SummEval, NewsRoom).}
\label{fig:eval-summ}
\end{figure}

\begin{figure}[!t]
\centering
\begin{tcolorbox}[width=1\textwidth, fontupper=\small, colback=blue!2, boxrule=0.9pt] 
\#\# Instructions

Please score on the \textcolor{purple}{[Aspect]} of a following natural language sentence generated according to a structured data expression.

Please return your score on how \textcolor{purple}{[Aspect]} is the sentence, in the scale of 1 to 6, with 1 being the lowest. \\

\#\# Example

\textcolor{blue}{[Data expression and text to be evaluated]} \\

\#\# Criteria for \textcolor{purple}{[Aspect]}

\textcolor{blue}{[Calibrated criteria for evaluating this aspect]} \\

\#\# Evaluation

Now, please evaluate how \textcolor{purple}{[Aspect]} is the is the generated sentence. (on a scale of 1-6, with 1 being the lowest)
Please carefully read the sentence and the structured data expression, and follow the scoring criteria to score the \textcolor{purple}{[Aspect]} of the sentence.

Please first return your score, and then provide your reasoning for the score.. \\

\textcolor{purple}{[Aspect]} Score (1-5):
\end{tcolorbox}
\caption{Prompt template for evaluation on data-to-text (SFRES, SFHOT).}
\label{fig:eval-d2t}
\end{figure}

\begin{figure}[!t]
\centering
\begin{tcolorbox}[width=1\textwidth, fontupper=\small, colback=blue!2, boxrule=0.9pt] Is the sentence supported by the article?

Answer 1 if the summary is factually consistent with the article, 0 otherwise.\\

\textcolor{blue}{[Article and Summary to be evaluated]} \\

\textcolor{blue}{[Calibrated criteria for evaluating hallucination]} \\

Answer:
\end{tcolorbox}
\caption{Prompt template for evaluation on evaluating hallucinations (QAGS-XSUM/CNN).}
\label{fig:eval-qags}
\end{figure}

\subsection{Criteria Refinement Templates}
\label{app:sr_templ}
\begin{figure}[t]
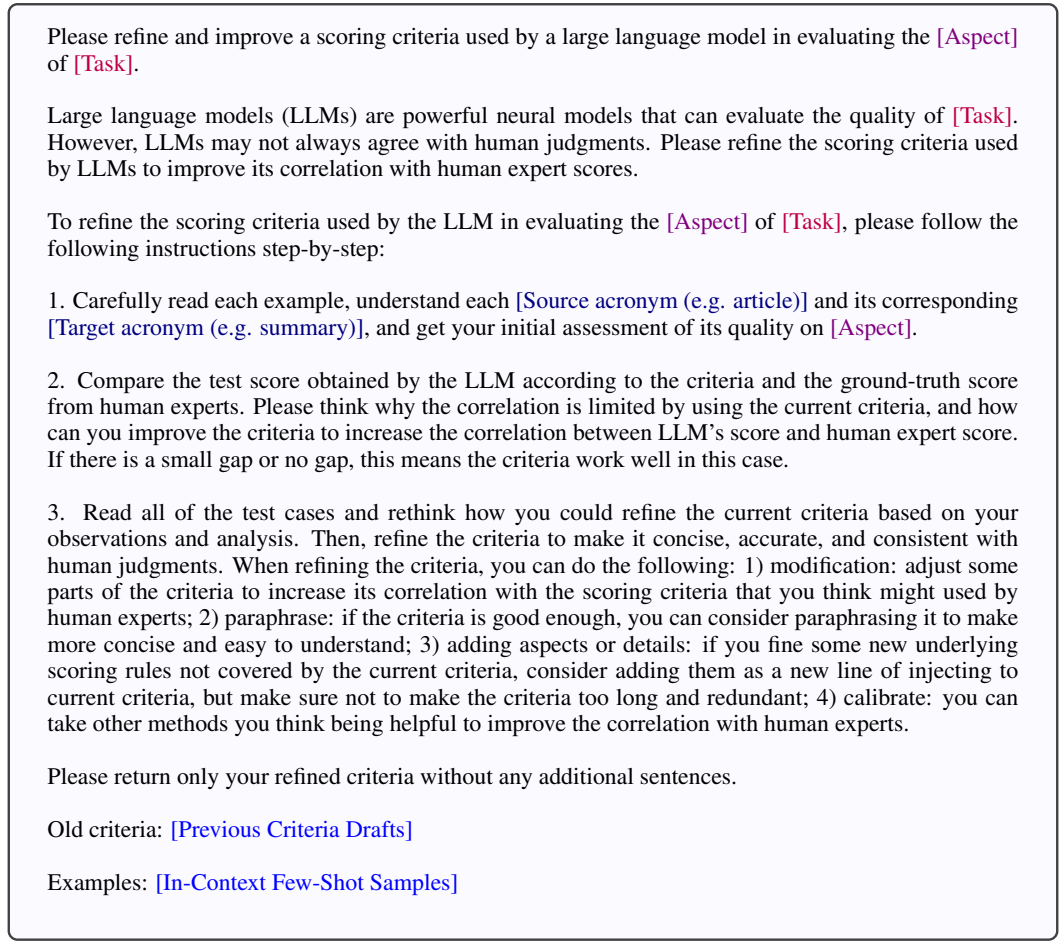

\centering
\begin{tcolorbox}[width=1\textwidth, fontupper=\small, colback=blue!2, boxrule=0.9pt] 
Please refine and improve a scoring criteria used by a large language model in evaluating the \textcolor{violet}{[Aspect]} of \textcolor{purple}{[Task]}. \\

Large language models (LLMs) are powerful neural models that can evaluate the quality of \textcolor{purple}{[Task]}.  However, LLMs may not always agree with human judgments. Please refine the scoring criteria used by LLMs to improve its correlation with human expert scores. \\

To refine the scoring criteria used by the LLM in evaluating the \textcolor{violet}{[Aspect]} of \textcolor{purple}{[Task]}, please follow the following instructions step-by-step: \\

1. Carefully read each example, understand each \textcolor{darkblue}{[Source acronym (e.g. article)]} and its corresponding \textcolor{darkblue}{[Target acronym (e.g. summary)]}, and get your initial assessment of its quality on \textcolor{violet}{[Aspect]}. \\

2. Compare the test score obtained by the LLM according to the criteria and the ground-truth score from human experts. Please think why the correlation is limited by using the current criteria, and how can you improve the criteria to increase the correlation between LLM's score and human expert score. If there is a small gap or no gap, this means the criteria work well in this case. \\

3. Read all of the test cases and rethink how you could refine the current criteria based on your observations and analysis. Then, refine the criteria to make it concise, accurate, and consistent with human judgments. When refining the criteria, you can do the following: 1) modification: adjust some parts of the criteria to increase its correlation with the scoring criteria that you think might used by human experts; 2) paraphrase: if the criteria is good enough, you can consider paraphrasing it to make more concise and easy to understand; 3) adding aspects or details: if you fine some new underlying scoring rules not covered by the current criteria, consider adding them as a new line of injecting to current criteria, but make sure not to make the criteria too long and redundant; 4) calibrate: you can take other methods you think being helpful to improve the correlation with human experts. \\

Please return only your refined criteria without any additional sentences. \\

Old criteria: \textcolor{blue}{[Previous Criteria Drafts]} \\

Examples: \textcolor{blue}{[In-Context Few-Shot Samples]} \\
\end{tcolorbox}
\caption{Prompt template for criteria refinement with GPT-4.}
\label{fig:sr-prompt}
\end{figure}

An example prompt template for criteria refinement can be found in Figure \ref{fig:sr-prompt}. As illustrated in the figure, we first fill in the aspect and tasks to the instructions, then prompt the LLM with the previous criteria, few-shot in-context samples of misaligned evaluations, together with suggested means of modifications to obtain a modified version of scoring criteria for this task.

\section{Extended Case Study}
\subsection{List of Criteria}
In this section, we present a case study on scoring criteria generated by \ourswb for each evaluation aspect of each benchmark throughout this study in Table \ref{tab:case-allcrit-nsr}, \ref{tab:case-allcrit-sme}, \ref{tab:case-allcrit-d2t} and \ref{tab:case-allcrit-qags}. As illustrated in the tables, scoring criteria generated with \ourswb are informative, covering significant rubrics to evaluate a given aspect of the target NLG task.

\begin{table*}[t]
\center \footnotesize
\tabcolsep0.1 in

\resizebox{1\textwidth}{!}{
\begin{tabular}{cp{0.9\textwidth}}
\toprule
\textbf{Aspect} & \textbf{Example Scoring Criteria} \\ \cmidrule(lr){1-2}

COH & \scriptsize - The summary should be a concise and accurate representation of the main points and ideas of the article. It should avoid including irrelevant or minor details that are not essential to the article's purpose or message. - The summary should have a clear and logical structure that follows the article's original order of information, or provides a coherent alternative order if it improves the summarization. The summary should use transitions and connectors to link the sentences and paragraphs coherently. - The summary should use the same or similar terminology and tone as the article, unless there is a need to simplify or clarify some terms for the intended audience. The summary should avoid introducing new or unfamiliar words or concepts that are not in the article or relevant to the summary. - The summary should maintain the same perspective and point of view as the article, unless there is a reason to shift or contrast it. The summary should not express the summarizer's own opinion, interpretation, or evaluation of the article, unless it is explicitly stated as such. - The summary should be grammatically correct and free of spelling, punctuation, and capitalization errors. The summary should use direct or indirect quotations and citations appropriately to acknowledge the source of the article. - The summary should be coherent and consistent with the article's topic and genre. The summary should avoid introducing information or claims that contradict or deviate from the article's main message. The summary should also avoid repeating information or using unnecessary filler words.\\ \cmidrule(lr){1-2}

INF & \scriptsize - A summary should convey the main idea and the most important details of the original article in a concise and clear way. - A summary should avoid repeating information that has already been mentioned or is irrelevant to the article's main point. - A summary should use accurate and specific words to describe the article's content, and avoid vague or ambiguous expressions. - A summary should maintain the same tone and perspective as the original article, and avoid adding personal opinions or interpretations. - A summary should follow the logical order and structure of the original article, and use transition words or phrases to connect sentences if needed. - Possible scoring scheme:  - 5: The summary meets all the criteria and has no significant flaws or errors.  - 4: The summary meets most of the criteria and has minor flaws or errors that do not affect the overall comprehension.  - 3: The summary meets some of the criteria and has moderate flaws or errors that affect the comprehension of some parts.  - 2: The summary meets few of the criteria and has major flaws or errors that affect the comprehension of most parts.  - 1: The summary meets none of the criteria and has severe flaws or errors that make it incomprehensible. \\ \cmidrule(lr){1-2}

FLU & \scriptsize - The score for fluency should reflect how well the summary is written in clear, natural and grammatical language, without unnecessary or confusing repetitions, contradictions or omissions. - A summary with a score of 5 should be fluent, coherent and engaging, with no errors or awkward expressions. It should use appropriate vocabulary, syntax and punctuation, and convey the main information and meaning of the article. - A summary with a score of 4 should be mostly fluent, coherent and engaging, with minor errors or awkward expressions that do not affect the overall meaning. It should use mostly appropriate vocabulary, syntax and punctuation, and convey most of the main information and meaning of the article. - A summary with a score of 3 should be somewhat fluent, coherent and engaging, but with some errors or awkward expressions that may affect the overall meaning or readability of the summary. It should use some appropriate vocabulary, syntax and punctuation, and convey some of the main information and meaning of the article, but may have some gaps or inaccuracies. - A summary with a score of 2 should be poorly fluent, coherent and engaging, with frequent errors or awkward expressions that significantly affect the overall meaning or readability of the summary. It should use limited or inappropriate vocabulary, syntax and punctuation, and convey little of the main information and meaning of the article, or may have some major distortions or misunderstandings. - A summary with a score of 1 should be very poorly fluent, coherent and engaging, with severe errors or awkward expressions that make the summary incomprehensible or unintelligible. It should use very limited or inappropriate vocabulary, syntax and punctuation, and convey none of the main information and meaning of the article, or may have some nonsensical or irrelevant content.
 \\ \cmidrule(lr){1-2}
REL & \scriptsize - The summary should capture the main topic, events, and outcomes of the article in a concise and accurate way. - The summary should not omit any essential information that is necessary to understand the article's purpose and significance. - The summary should not include any irrelevant or redundant details that distract from the article's main points or introduce confusion. - The summary should use the same or similar terminology and tone as the article, unless the article uses obscure or jargon words that need to be simplified. - The summary should reflect the article's structure and organization, presenting the information in a logical and coherent order. Examples of scoring: - Score 5: The summary meets all the criteria for relevance and provides a clear and comprehensive overview of the article, without any errors or gaps. - Score 4: The summary meets most of the criteria for relevance and provides a mostly clear and comprehensive overview of the article, but may have some minor errors or gaps, such as missing a minor detail, using a slightly different word, or omitting a transition. - Score 3: The summary meets some of the criteria for relevance and provides a partially clear and comprehensive overview of the article, but has some noticeable errors or gaps, such as missing a key detail, using a vague or inaccurate word, or skipping a logical connection. - Score 2: The summary meets few of the criteria for relevance and provides a vaguely clear and comprehensive overview of the article, but has many errors or gaps, such as missing several important details, using inappropriate or misleading words, or presenting the information in a confusing or contradictory order. - Score 1: The summary meets none or almost none of the criteria for relevance and provides a unclear and incomplete overview of the article, with severe errors or gaps, such as missing the main topic, using incorrect or irrelevant words, or omitting the entire conclusion.	\\

\bottomrule
\end{tabular}
}

\caption{\textbf{Case study on criteria} on each aspect for NewsRoom generated by \ours.}
\label{tab:case-allcrit-nsr}
\end{table*}

\begin{table*}[t]
\center \footnotesize
\tabcolsep0.1 in

\resizebox{1\textwidth}{!}{
\begin{tabular}{cp{0.9\textwidth}}
\toprule
\textbf{Aspect} & \textbf{Example Scoring Criteria} \\ \cmidrule(lr){1-2}

COH & \scriptsize Coherence is the quality of being consistent, logical, and well-organized in the summary. A summary is coherent if it accurately captures the main ideas and key information from the article, and presents them in a clear and concise manner. A summary is not coherent if it omits important details, contradicts the article, or introduces irrelevant or confusing information. The score for coherence is based on the following scale: - 5: The summary is very coherent, with no errors or flaws. - 4: The summary is mostly coherent, with only minor errors or gaps. - 3: The summary is somewhat coherent, but has some significant errors or omissions. - 2: The summary is poorly coherent, with many errors, inconsistencies, or redundancies. - 1: The summary is not coherent at all, with little or no relation to the article. \\ \cmidrule(lr){1-2}

CON & \scriptsize - A summary is consistent with the article if it accurately and faithfully reflects the main points, facts, and tone of the article without changing, adding, or omitting any significant information. - A summary should avoid introducing any errors, contradictions, or distortions of the original article, unless they are explicitly marked as the summary writer's opinions or interpretations. - A summary should use clear and precise language that matches the style and genre of the article, and avoid any vague or ambiguous expressions that could mislead the reader or obscure the meaning of the article. - A summary should maintain the logical structure and coherence of the article, and present the information in a well-organized and easy-to-follow manner. - A summary should be concise and avoid any unnecessary or redundant details that do not contribute to the main purpose or message of the article. \\ \cmidrule(lr){1-2}

FLU & \scriptsize - A fluent summary should reflect the main content and structure of the original article, using clear and coherent language that avoids redundancy and errors. - A fluent summary should retain the key information and details from the article, without introducing any irrelevant or inaccurate information that distorts the meaning of the original text. - A fluent summary should use appropriate transition words, connectors, and referents to ensure the logical flow and cohesion of the summary, and avoid abrupt or confusing shifts in topic or perspective. - A fluent summary should use varied and precise vocabulary and grammar that suits the tone and style of the article, and avoid repetition or ambiguity. - A fluent summary should use correct spelling, punctuation, and capitalization throughout the summary, and follow the conventions of standard written English. A possible scoring rubric based on these criteria is: - 5: The summary is fluent and meets all the criteria listed above. It captures the main points and details of the article accurately and effectively, using clear and coherent language that follows the logical structure of the article. The summary uses appropriate transition words, connectors, and referents to ensure cohesion, and varied and precise vocabulary and grammar that suits the tone and style of the article. The summary has no or minimal errors in spelling, punctuation, and capitalization. - 4: The summary is mostly fluent and meets most of the criteria listed above. It captures the main points and details of the article fairly well, using mostly clear and coherent language that follows the logical structure of the article. The summary uses mostly appropriate transition words, connectors, and referents to ensure cohesion, and mostly varied and precise vocabulary and grammar that suits the tone and style of the article. The summary has few errors in spelling, punctuation, and capitalization. - 3: The summary is somewhat fluent and meets some of the criteria listed above. It captures some of the main points and details of the article, but may omit or misrepresent some important information. The summary uses somewhat clear and coherent language, but may deviate from the logical structure of the article or have some lapses in cohesion. The summary uses some appropriate transition words, connectors, and referents, but may also have some inappropriate or confusing ones. The summary uses some varied and precise vocabulary and grammar, but may also have some repetition or ambiguity. The summary has several errors in spelling, punctuation, and capitalization. - 2: The summary is not very fluent and meets few of the criteria listed above. It captures few of the main points and details of the article, and may omit or misrepresent many important information. The summary uses unclear or incoherent language, and does not follow the logical structure of the article or have much cohesion. The summary uses few or no appropriate transition words, connectors, and referents, and may have many inappropriate or confusing ones. The summary uses limited or imprecise vocabulary and grammar, and may have many repetition or ambiguity. The summary has many errors in spelling, punctuation, and capitalization. - 1: The summary is not fluent and meets none of the criteria listed above. It captures none or almost none of the main points and details of the article, and may omit or misrepresent most or all of the important information. The summary uses incomprehensible or irrelevant language, and does not follow the logical structure of the article or have any cohesion. The summary uses no or almost no appropriate transition words, connectors, and referents, and may have only inappropriate or confusing ones. The summary uses very limited or inaccurate vocabulary and grammar, and may have only repetition or ambiguity. The summary has numerous and severe errors in spelling, punctuation, and capitalization. \\ \cmidrule(lr){1-2}
REL & \scriptsize - A summary is relevant if it captures the main points or the most important information from the article, without leaving out any crucial details or adding any unnecessary or inaccurate ones. - A summary is more relevant if it uses the same or similar terms and expressions as the article, as long as they are clear and concise. - A summary is less relevant if it omits or misrepresents some of the key facts or arguments from the article, or if it introduces irrelevant or erroneous information that is not supported by the article. - A summary is irrelevant if it does not correspond to the article at all, or if it only mentions a minor or peripheral aspect of the article. \\

\bottomrule
\end{tabular}
}

\caption{\textbf{Case study on criteria} on each aspect for SummEval generated by \ours.}
\label{tab:case-allcrit-sme}
\end{table*}

\begin{table*}[t]
\center \footnotesize
\tabcolsep0.1 in

\resizebox{1\textwidth}{!}{
\begin{tabular}{cp{0.9\textwidth}}
\toprule
\textbf{Aspect} & \textbf{Example Scoring Criteria} \\ \cmidrule(lr){1-2}

INF & \scriptsize - A natural language sentence is informative if it conveys all the relevant information from the data expression, without omitting, adding, or distorting any facts. - A sentence is more informative if it uses clear and natural language, without grammatical errors, ambiguity, or redundancy. - A sentence is less informative if it leaves out some information from the data expression, or if it uses vague, unnatural, or incorrect language. - A possible scoring rule for informativeness is as follows: - 6: The sentence conveys all the information from the data expression, using clear and natural language. - 5.5: The sentence conveys all the information from the data expression, using mostly clear and natural language, but with minor issues (e.g., word choice, punctuation, etc.) - 5: The sentence conveys all the information from the data expression, but with some issues in language clarity or naturalness. - 4.5: The sentence conveys most of the information from the data expression, using clear and natural language, but omitting one detail. - 4: The sentence conveys most of the information from the data expression, but with some issues in language clarity or naturalness, or omitting more than one detail. - 3: The sentence conveys some of the information from the data expression, but with significant issues in language clarity or naturalness, or omitting several details. - 2: The sentence conveys little of the information from the data expression, or with major issues in language clarity or naturalness, or adding or distorting facts. - 1: The sentence conveys none of the information from the data expression, or with unintelligible or irrelevant language.	\\ \cmidrule(lr){1-2}

NAT & \scriptsize	- A natural language sentence is natural if it is fluent, coherent, grammatical, and conveys the meaning of the data expression accurately and concisely. - The score of naturalness ranges from 1 to 6, where 1 is the lowest and 6 is the highest. - The score is assigned based on the following criteria: - A sentence that is completely natural, without any errors or awkwardness, and expresses the data expression fully and succinctly, gets a 6.  - A sentence that is mostly natural, with minor errors or redundancy, and expresses the data expression adequately, gets a 5 or 5.5.  - A sentence that is somewhat natural, with noticeable errors or incompleteness, and expresses the data expression partially or vaguely, gets a 4 or 4.5.  - A sentence that is barely natural, with serious errors or confusion, and expresses the data expression incorrectly or irrelevantly, gets a 3 or 3.5.  - A sentence that is not natural at all, with unacceptable errors or nonsense, and does not express the data expression at all, gets a 1 or 2.\\

\bottomrule
\end{tabular}
}

\caption{\textbf{Case study on criteria} on each aspect for SFRES generated by \ours.}
\label{tab:case-allcrit-d2t}
\end{table*}

\begin{table*}[t]
\center \footnotesize
\tabcolsep0.1 in

\resizebox{1\textwidth}{!}{
\begin{tabular}{cp{0.9\textwidth}}
\toprule
\textbf{Aspect} & \textbf{Example Scoring Criteria} \\ \cmidrule(lr){1-2}

FACT & \scriptsize - Score 1 if the sentence accurately and concisely summarizes the main facts and information from the article, without omitting, distorting, or adding any significant details. - Score 0.75 if the sentence summarizes the main facts and information from the article, but has some minor issues such as: omitting, distorting, or adding some less important details; using vague or imprecise language; or being too long or verbose. - Score 0.5 if the sentence captures some of the facts and information from the article, but has some major issues such as: omitting, distorting, or adding some important details; using incorrect or misleading language; or being too short or incomplete. - Score 0.25 if the sentence only captures a few of the facts and information from the article, and has many issues such as: omitting, distorting, or adding most of the details; using irrelevant or contradictory language; or being too general or specific.- Score 0 if the sentence does not capture any of the facts and information from the article, or contradicts or misrepresents the article entirely. \\

\bottomrule
\end{tabular}
}

\caption{\textbf{Case study on criteria} on each aspect for QAGS-CNN generated by \ours.}
\label{tab:case-allcrit-qags}
\end{table*}

\end{document}